\pdfoutput=1

\documentclass[11pt]{article} 

\usepackage{acl}

\usepackage{times}
\usepackage{latexsym}

\usepackage[T1]{fontenc}

\usepackage[utf8]{inputenc}

\usepackage{microtype}
\usepackage{inconsolata}

\usepackage{makecell}
\usepackage{graphicx}
\usepackage{multirow}
\usepackage{utfsym}
\usepackage{amsthm,amsmath,amssymb}
\usepackage{xspace}
\usepackage{booktabs}
\usepackage{soul}
\usepackage{color, xcolor}

\usepackage{subfigure}
\usepackage{pifont}
\usepackage{enumitem}
\setenumerate[1]{itemsep=0pt,parsep=0pt,topsep=0pt}
\setitemize[1]{itemsep=5pt,parsep=0pt,topsep=5pt,leftmargin=8pt}
\newcommand{\benchmark}{\texttt{PAGED}\xspace}
\soulregister{\cite}7 
\soulregister{\citep}7 
\soulregister{\citet}7 
\soulregister{\ref}7 
\soulregister{\pageref}7 

\title{\textbf{\benchmark}: A Benchmark for Procedural Graphs Extraction from Documents}

\author{
{Weihong Du\textsuperscript{1}\textsuperscript{2}\quad Wenrui Liao\textsuperscript{1}\textsuperscript{2}\quad Hongru Liang\textsuperscript{1}\textsuperscript{2}\thanks{\quad Corresponding author}\quad Wenqiang Lei\textsuperscript{1}\textsuperscript{2}}
\\
{\textsuperscript{1}College of Computer Science, Sichuan University, China} \\
{\textsuperscript{2}Engineering Research Center of Machine Learning and Industry Intelligence,} \\
{Ministry of Education, China} \\
\texttt{\{duweihong, liaowenrui\}@stu.scu.edu.cn}\\
\texttt{\{lianghongru, wenqianglei\}@scu.edu.cn}
}

\begin{document}

\maketitle

\begin{abstract}

Automatic extraction of procedural graphs from documents creates a low-cost way for users to easily understand a complex procedure by skimming visual graphs. Despite the progress in recent studies, it remains unanswered: \textit{whether the existing studies have well solved this task}~(\textit{Q1}) and \textit{whether the emerging large language models~(LLMs) can bring new opportunities to this task}~(\textit{Q2}). To this end, we propose a new benchmark \benchmark, equipped with a large high-quality dataset and standard evaluations. It investigates five state-of-the-art baselines, revealing that they fail to extract optimal procedural graphs well because of their heavy reliance on hand-written rules and limited available data. We further involve three advanced LLMs in \benchmark and enhance them with a novel self-refine strategy. The results point out the advantages of LLMs in identifying textual elements and their gaps in building logical structures. 
We hope \benchmark can serve as a major landmark for automatic procedural graph extraction and the investigations in \benchmark can provide valuable insights into the research on logical reasoning among non-sequential elements.
The code and dataset are available in \href{https://github.com/SCUNLP/PAGED}{https://github.com/SCUNLP/PAGED}. 
\end{abstract}
\section{Introduction}
\label{sec:intro}
\begin{figure*}[t]
    \centering    
    \subfigure[Procedural Graph]{\includegraphics[width=1\textwidth]{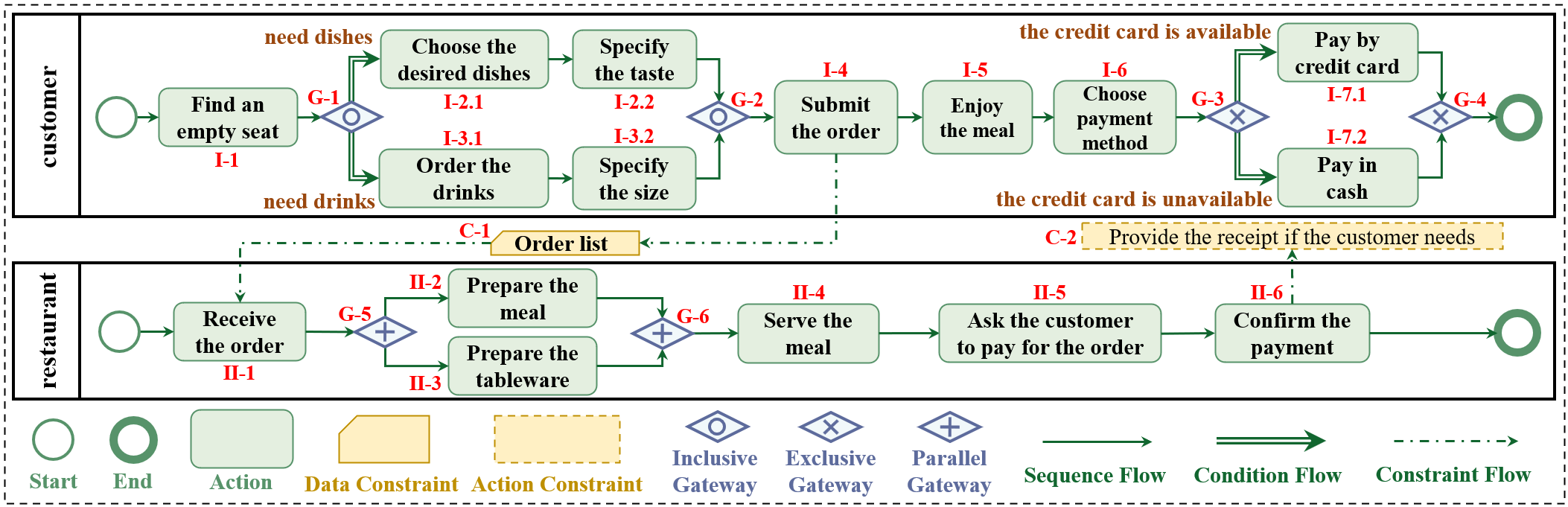}\label{fig:task_b}}\vspace{-5pt}    
    
    \subfigure[Procedural Document]{\includegraphics[width=1\textwidth]{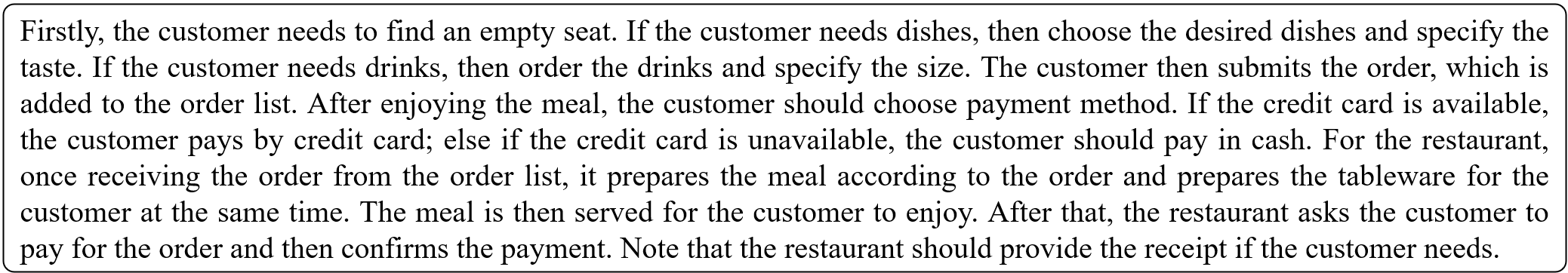}\label{fig:task_a}}
    
    \caption{The procedure of how a restaurant serves the customers in procedural graph~(a) and document~(b).}
    \label{fig:task}
\end{figure*}

Procedural graphs, though can intuitively represent the execution of actions for goal achievement~\cite{momouchi1980control, ren2023constructing}, suffer from the high cost of expert-construction~\cite{herbst1999inductive, maqbool2019comprehensive}. 
The automatic extraction of procedural graphs from procedural documents thus has huge potential, as it would enable users to easily understand how to logically perform a goal~(e.g, how a restaurant serves the customers) by skimming visual graphs~(e.g., Figure~\ref{fig:task_b} instead of reading lengthy documents~(e.g., Figure~\ref{fig:task_a}). 

However, obtaining optimal procedural graphs is not easy --- as shown in Figure~\ref{fig:task_b}, it requires representing not only sequential actions in the procedure~(e.g., \uppercase\expandafter{\romannumeral1}-4  $\rightarrow$ \uppercase\expandafter{\romannumeral1}-5), but also non-sequentially executed actions~(e.g., \uppercase\expandafter{\romannumeral1}-7.1 \& \uppercase\expandafter{\romannumeral1}-7.2]) and vital constraints for the actions~(e.g., C-2).  
Off-the-shelf attempts only meet part of the requirements. For example, \citet{bellan2023pet, ren2023constructing} fail to represent ``customers can order only the dishes or drinks, or both''~(\uppercase\expandafter{\romannumeral1}-2.1 \& \uppercase\expandafter{\romannumeral1}-2.2) due to its inherent limitation for representing complex non-sequential actions. 
Besides, current solutions only focus on hand-written rules~\cite{sholiq2022generating} or customized networks~\cite{bellan2023pet} on a small group of cherry-picked instances. This brings up the question of \textit{whether the existing studies have well solved the automated extraction of procedural graphs from procedural documents}~(\textit{Q1}). If the answer is ``no'', we are also interested in \textit{whether the emerging large language models~(LLMs) can bring new opportunities to this task}~(\textit{Q2}). 

To answer \textit{Q1}, we propose to construct a standard benchmark for the \underline{P}rocedur\underline{A}l \underline{G}raphs \underline{E}xtraction from \underline{D}ocuments~(\benchmark). As far as we know, there lack of large-scale datasets of document-graph pairs for training and evaluating optimal procedural graph extraction models~(cf., Table~\ref{datasets}). We want to equip \benchmark with the largest publicly available dataset. Although there exist plenty of procedural documents on the Internet, it is too costly to filter the low-quality ones and annotate optimal procedural graphs matching all requirements. As a remedy, we build the dataset based on a model collection of business process~\cite{dumas2018fundamentals}, which has summarized business processes into high-quality procedural graphs with complete sequential actions, non-sequential actions, and constraints. Thus, constructing procedural document-graph pairs turns into assigning a suitable procedural document to a given procedural graph. We approach it via a three-stage pipeline --- we progressively transfer the structured information on the procedural graph into natural language text, adjust its narration, and improve the coherence and naturalness, finally generating a suitable document. In this way, we develop a dataset with 3,394 high-quality procedural document-graph pairs that are about ten times larger than the previous largest datasets~\cite{ackermann2021data,qian2020approach}. According to the underlying structure of optimal procedural graphs, we introduce {three metrics} to evaluate five state-of-the-art methods~\cite{sonbol2023machine,neuberger2023beyond,sholiq2022generating}. 

To further answer \textit{Q2}, we investigate the performance of three advanced LLMs~(Flan-T5~\cite{chung2022scaling}, ChatGPT~\cite{ouyang2022training} and Llama2~\cite{touvron2023llama}) and utilize a self-refine strategy to improve the ability of LLMs. In total, we evaluate ten methods in our \benchmark benchmark. Extensive experiments on our benchmark reveal that existing studies struggle to accurately extract sequential actions, constraints, and organize non-sequential actions of procedural documents. While LLMs have shown significant improvement in sequential action and constraint extraction, they still face challenges with non-sequential action organization. Our detailed analysis of the results leads us to propose improvement strategies to help large language models better understand non-sequential actions and use correct gateways to represent them. We hope \benchmark can be a key milestone for automatic procedural graphs extraction, offering insights into research on logical reasoning among non-sequential elements.
In summary, we highlight \benchmark as follows:

\begin{itemize}
    \item We build a novel benchmark named \benchmark, which standardly evaluates the progress of current procedural graphs extraction from documents and explores the potential of emerging LLMs.
    \item We equip \benchmark with the largest procedural document-graph dataset, whose high quality is achieved by a three-stage pipeline and verified via both automatic and human evaluation. 
    \item We systematically evaluate state-of-the-art solutions in \benchmark and reveal that they have trouble extracting optimal procedural graphs due to their heavy reliance on hand-written rules and limited available data. 
    \item We investigate advanced LLMs in \benchmark and empower them with a self-refine strategy, showing their adavantages in identifying sequential actions and constraints, and pointing out their gaps in building complex logic of graphs. 
\end{itemize}

\begin{table*}[t]
    \caption{Comparisons between our dataset with the existing datasets.}
    \label{datasets}
    \centering
    \scalebox{0.75}{
        \begin{tabular}{|c|c|c|ccc|cc|c|}
            \hline
                                         &                                     &                                              & \multicolumn{3}{c|}{Non-sequential Actions}                       & \multicolumn{2}{c|}{Constraints}                                  &                                                                                                                                                                                                                \\ \cline{4-8}
            \multirow{-2}{*}{Dataset}    & \multirow{-2}{*}{ \makecell{Samples                                                                                                                                                                                                                                                                                                                                                                                                         \\ Num} } & \multirow{-2}{*}{ \makecell{Sequential \\ Actions} } & \multicolumn{1}{c|}{ \makecell{Exclusive \\ Gateway} }                 & \multicolumn{1}{c|}{\makecell{Inclusive \\ Gateway}}                 & \makecell{Parallel \\ Gateway}                  & \multicolumn{1}{c|}{\makecell{Data \\ Constraint}}                   & \makecell{Action \\ Constraint}                 & \multirow{-2}{*}{\makecell{Publicly \\ Available}} \\ \hline
            \citet{friedrich2011process}  & 47                                  & {\color[HTML]{009901} \usym{1F5F8}}          & \multicolumn{1}{c|}{{\color[HTML]{009901} \usym{1F5F8}}}          & \multicolumn{1}{c|}{{\color[HTML]{FE0000} \usym{2715} }}          & {\color[HTML]{009901} \usym{1F5F8}}          & \multicolumn{1}{c|}{{\color[HTML]{009901} \usym{1F5F8}}}          & {\color[HTML]{FE0000} \usym{2715} }          & {\color[HTML]{009901} \usym{1F5F8}}          \\ \hline
            \citet{epure2015automatic}    & 34                                  & {\color[HTML]{009901} \usym{1F5F8} }         & \multicolumn{1}{c|}{{\color[HTML]{009901} \usym{1F5F8}}}          & \multicolumn{1}{c|}{{\color[HTML]{FE0000} \usym{2715} }}          & {\color[HTML]{009901} \usym{1F5F8}}          & \multicolumn{1}{c|}{{\color[HTML]{FE0000} \usym{2715} }}          & {\color[HTML]{FE0000} \usym{2715} }          & {\color[HTML]{FE0000} \usym{2715}}           \\ \hline
            \citet{ferreira2017semi}      & 56                                  & {\color[HTML]{009901} \usym{1F5F8} }         & \multicolumn{1}{c|}{{\color[HTML]{009901} \usym{1F5F8}}}          & \multicolumn{1}{c|}{{\color[HTML]{FE0000} \usym{2715} }}          & {\color[HTML]{009901} \usym{1F5F8}}          & \multicolumn{1}{c|}{{\color[HTML]{FE0000} \usym{2715} }}          & {\color[HTML]{FE0000} \usym{2715} }          & {\color[HTML]{FE0000} \usym{2715}}           \\ \hline
            \citet{mendling2019natural}   & 103                                 & {\color[HTML]{009901} \usym{1F5F8}}          & \multicolumn{1}{c|}{{\color[HTML]{FE0000} \usym{2715} }}          & \multicolumn{1}{c|}{{\color[HTML]{FE0000} \usym{2715} }}          & {\color[HTML]{FE0000} \usym{2715} }          & \multicolumn{1}{c|}{{\color[HTML]{FE0000} \usym{2715} }}          & {\color[HTML]{FE0000} \usym{2715} }          & {\color[HTML]{009901} \usym{1F5F8}}          \\ \hline
            \citet{quishpi2020extracting} & 121                                 & {\color[HTML]{FE0000} \usym{2715} }          & \multicolumn{1}{c|}{{\color[HTML]{FE0000} \usym{2715} }}          & \multicolumn{1}{c|}{{\color[HTML]{FE0000} \usym{2715} }}          & {\color[HTML]{FE0000} \usym{2715} }          & \multicolumn{1}{c|}{{\color[HTML]{FE0000} \usym{2715} }}          & {\color[HTML]{FE0000} \usym{2715} }          & Partial                                      \\ \hline
            \citet{qian2020approach}      & 360                                 & {\color[HTML]{FE0000} \usym{2715}}           & \multicolumn{1}{c|}{{\color[HTML]{FE0000} \usym{2715} }}          & \multicolumn{1}{c|}{{\color[HTML]{FE0000} \usym{2715} }}          & {\color[HTML]{FE0000} \usym{2715} }          & \multicolumn{1}{c|}{{\color[HTML]{FE0000} \usym{2715} }}          & {\color[HTML]{FE0000} \usym{2715} }          & {\color[HTML]{009901} \usym{1F5F8} }         \\ \hline
            \citet{ackermann2021data}     & 358                                 & {\color[HTML]{FE0000} \usym{2715}}           & \multicolumn{1}{c|}{{\color[HTML]{FE0000} \usym{2715} }}          & \multicolumn{1}{c|}{{\color[HTML]{FE0000} \usym{2715} }}          & {\color[HTML]{FE0000} \usym{2715} }          & \multicolumn{1}{c|}{{\color[HTML]{FE0000} \usym{2715} }}          & {\color[HTML]{FE0000} \usym{2715} }          & {\color[HTML]{009901} \usym{1F5F8} }         \\ \hline
            \citet{lopez2021declarative}  & 37                                  & {\color[HTML]{009901} \usym{1F5F8}}          & \multicolumn{1}{c|}{{\color[HTML]{FE0000} \usym{2715} }}          & \multicolumn{1}{c|}{{\color[HTML]{FE0000} \usym{2715} }}          & {\color[HTML]{FE0000} \usym{2715} }          & \multicolumn{1}{c|}{{\color[HTML]{FE0000} \usym{2715} }}          & {\color[HTML]{FE0000} \usym{2715} }          & Partial                                      \\ \hline
            \citet{bellan2023pet}         & 45                                  & {\color[HTML]{009901} \usym{1F5F8}}          & \multicolumn{1}{c|}{{\color[HTML]{009901} \usym{1F5F8}}}          & \multicolumn{1}{c|}{{\color[HTML]{FE0000} \usym{2715} }}          & {\color[HTML]{009901} \usym{1F5F8}}          & \multicolumn{1}{c|}{{\color[HTML]{FE0000} \usym{2715} }}          & {\color[HTML]{FE0000} \usym{2715} }          & {\color[HTML]{009901} \usym{1F5F8}}          \\ \hline
            \citet{liang2023knowing}      & 200                                 & {\color[HTML]{009901} \usym{1F5F8}}          & \multicolumn{1}{c|}{{\color[HTML]{FE0000} \usym{2715} }}          & \multicolumn{1}{c|}{{\color[HTML]{FE0000} \usym{2715} }}          & {\color[HTML]{FE0000} \usym{2715} }          & \multicolumn{1}{c|}{{\color[HTML]{FE0000} \usym{2715} }}          & {\color[HTML]{FE0000} \usym{2715} }          & {\color[HTML]{009901} \usym{1F5F8}}          \\ \hline
            \citet{ren2023constructing}   & 283                                 & {\color[HTML]{009901} \usym{1F5F8}}          & \multicolumn{1}{c|}{{\color[HTML]{FE0000} \usym{2715} }}          & \multicolumn{1}{c|}{{\color[HTML]{FE0000} \usym{2715} }}          & {\color[HTML]{FE0000} \usym{2715} }          & \multicolumn{1}{c|}{{\color[HTML]{FE0000} \usym{2715} }}          & {\color[HTML]{FE0000} \usym{2715} }          & {\color[HTML]{009901} \usym{1F5F8}}          \\ \hline
            \textbf{ours}                & \textbf{3,394}                       & {\color[HTML]{009901} \textbf{\usym{1F5F8}}} & \multicolumn{1}{c|}{{\color[HTML]{009901} \textbf{\usym{1F5F8}}}} & \multicolumn{1}{c|}{{\color[HTML]{009901} \textbf{\usym{1F5F8}}}} & {\color[HTML]{009901} \textbf{\usym{1F5F8}}} & \multicolumn{1}{c|}{{\color[HTML]{009901} \textbf{\usym{1F5F8}}}} & {\color[HTML]{009901} \textbf{\usym{1F5F8}}} & {\color[HTML]{009901} \textbf{\usym{1F5F8}}} \\ \hline
        \end{tabular}
    }
\vspace{10pt}
\end{table*}

\section{Related Work}

\paragraph{Procedural Graph Extraction} 
Existing studies only meet part of the requirements for optimal procedural graph extraction. 
Earlier studies mainly focus on extracting sequential actions~\cite{pal2021constructing, lopez2021declarative, ren2023constructing}. 
While \citet{epure2015automatic, honkisz2018concept, bellan2023pet} explore non-sequential actions, they do not cover scenarios like \uppercase\expandafter{\romannumeral1}-2.1 \& \uppercase\expandafter{\romannumeral1}-2.2 in Figure~\ref{fig:task_b}. \citet{friedrich2011process} aids in data constraint extraction but overlooks action constraints. 
Besides, current studies heavily rely on hand-written rules and templates~\cite{epure2015automatic, honkisz2018concept}, resulting in poor generalization. \citet{bellan2023pet} trains a neural network model but only learns 45 samples, whose effectiveness remains questionable. 
Hence, we propose to construct a standard benchmark to reveal the performance of existing studies and highlight the challenges for the extraction of optimal procedural graphs.

\paragraph{Datasets}
As shown in Table~\ref{datasets}, current datasets consist of a small group of cherry-picked instances. 
Some datasets are not publicly available~\cite{epure2015automatic, ferreira2017semi}. 
Some datasets only focus on sentence-level actions extraction, lacking both sequential and non-sequential actions~\cite{quishpi2020extracting, qian2020approach, ackermann2021data}. While other studies contain sequential actions~\citet{friedrich2011process, mendling2019natural, lopez2021declarative, bellan2023pet, liang2023knowing, ren2023constructing}, they do not cover all types of non-sequential actions. Moreover, almost all existing datasets ignore vital constraints related to the actions in procedural graphs. To this end, we construct a new procedural document-graph dataset that is nearly ten times larger than the previous largest datasets~\cite{ackermann2021data,qian2020approach}. Each sample consists of a high-quality procedural document and its procedural graph with complete sequential actions, non-sequential actions, and constraints.

\paragraph{Data2Text}
Data2Text task aims at transferring structured data such as graphs into natural language text~\cite{duong2023learning, lin2023survey}. 
Current studies~\cite{su2021few, kasner2022neural} only focus on the transformation of factual knowledge --- knowledge about features of things, making it difficult to deal with procedural knowledge --- execution of sequential and non-sequential actions in the procedural graphs. 
Moreover, current studies can only manage discrete components~\cite{ye2019variational, fu2020partially}, while extracting procedural graphs requires handling complex logic of sequential action, non-sequential actions and their constraints. In this paper, we propose a three-stage pipeline to bridge the gap between complex graphs and lengthy documents, ensure logical descriptions of generated documents, and solve the issues of fluency and coherence in generated documents.

\paragraph{Large Language Models}
The emerging LLMs have presented competitive results in a wide range of tasks~\cite{zhao2023survey}, but are barely used for procedural graph extraction. 
The only exception is \citet{bellan2022leveraging}, which makes a shallow attempt to extract sequential actions and deal with partial non-sequential actions with LLMs, and performs poorly for gateway extraction. It remains unanswered whether LLMs' ability to understand the inherent structure of long contexts can improve the procedural graphs extraction from documents. To this end, we involve Flan-T5~\cite{chung2022scaling}, ChatGPT~\cite{ouyang2022training} and Llama2~\cite{touvron2023llama} in our \benchmark and design a self-refine strategy to demonstrate the opportunities and gaps of LLMs in this task. We hope this can help to explore more possibilities that LLMs bring to this field.

\section{Dataset}
\label{sec:data}
It is too costly to conduct an expert annotation of optimal procedural graphs for a large number of documents. To this end, we build our dataset upon a model collection of business process~\cite{dumas2018fundamentals}, which has defined optimal procedural graphs covering the whole business process management lifecycle. The dataset construction turns into a data2text task --- generates suitable documents for given procedural graphs.

\subsection{Preliminary}
Figure~\ref{fig:task_b} presents an example of the optimal procedural graph. It describes the procedure of how a restaurant serves customers and involves two actors~(customer and restaurant). 
Each actor starts from the ``Start'' node, carries out actions following the logic in the graph, and ends at the ``End'' node. If the actions are performed sequentially, they are connected by the ``Sequence Flow''. Otherwise, there is an ``Inclusive Gateway'', ``Exclusive Gateway'' or ``Parallel Gateway'' indicating the following actions are non-sequential ones. Both the ``Inclusive Gateway''~(G-1) and ``Exclusive Gateway''~(G-3) mean that the following action is performed under the condition on the connected ``Condition Flow''. The difference is that there is one and only one condition after the ``Exclusive Gateway'' can be met, while this does not apply to the ``Inclusive Gateway''. The ``Parallel Gateway''~(G-5) represents that the following actions are performed in parallel. Note that, all gateways appear in pairs and the latter ones~(G-4, G-2, G-6) indicate the end of the non-sequential activities. Additionally, the ``Data Constraint'' and ``Action Constraint'' represent the necessary data~(C-1) and essential notices~(C-2) for actions connected by the ``Constraint Flow'', respectively.

\subsection{Dataset Construction}
With these high-quality procedural graphs, we then perform the dataset construction as a data2text task~\cite{lin2023survey}. Specifically, {we design a three-stage pipeline: 1) Decomposition \& Transformation that decomposes the graph into fragmented spans/sentences in natural language; 2) Grouping \& Ordering that logically organizes the procedural fragments; 3) Aggregating \& Smoothing that unifies the fragments into high-quality documents}.

\begin{figure}[t]
    \centering
    \includegraphics[width=1\linewidth]{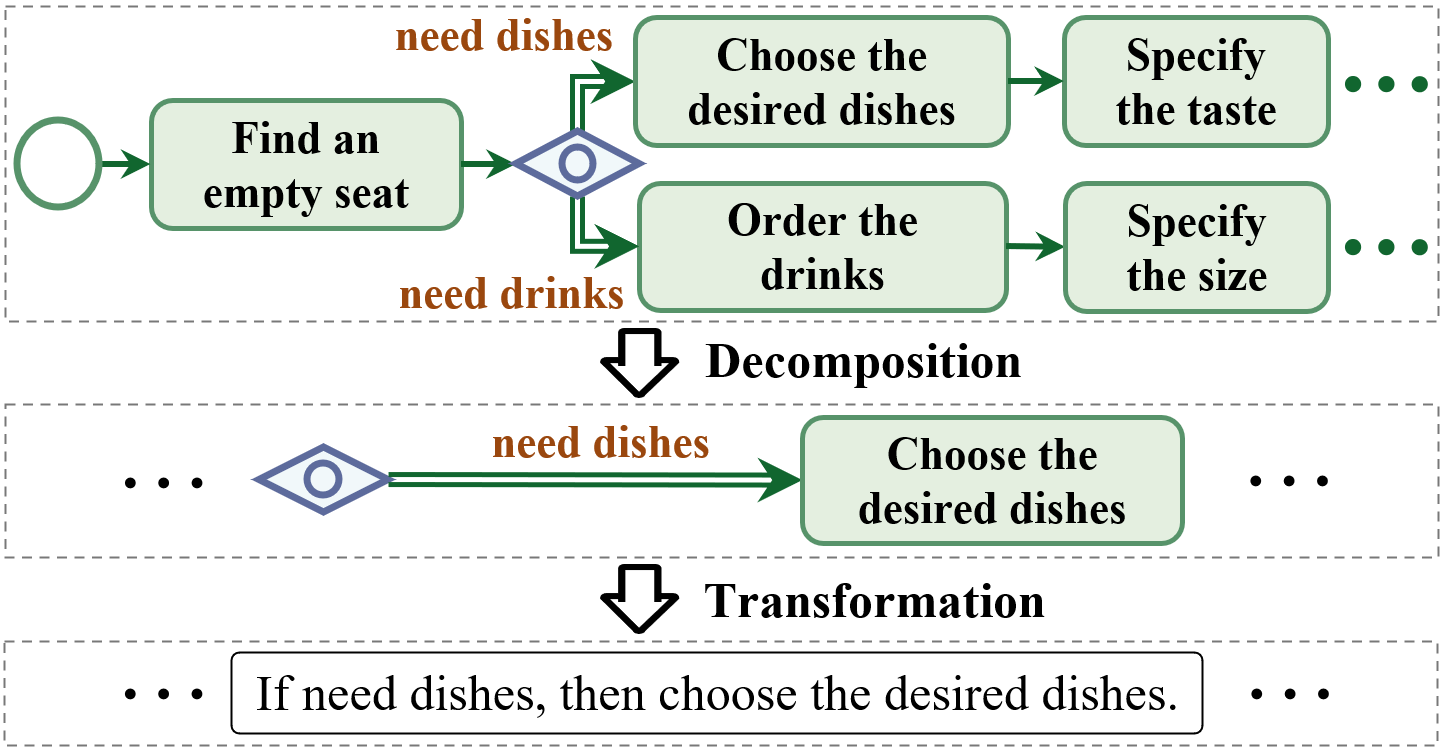}
    \caption{Illustration of decomposing the graph into units and transforming a unit into a procedural fragment.
    }
    \label{fig:Decomposition}
\end{figure}

\paragraph{Decomposition \& Transformation}
To narrow the huge gap between the complex graph and the length document, we first decompose the graph into minimal meaningful units. We define a vocabulary with nineteen units, each of which consists of actions, gateways or constraints connected by flow~(cf., Appx.~\ref{app:DT}). We also design nineteen hand-written templates for the vocabulary. Based on the unit vocabulary, we perform the breadth-first search strategy~\cite{10.1016/0004-3702(85)90084-0} over the graph, as shown in Figure~\ref{fig:Decomposition}. Particularly, some actions may be repeatedly walked to preserve the sequential execution relation between adjacent actions in the graph. For example, ``Order drinks'' is walked twice, forming ``If need drinks, then order drinks.'' and ``Order drinks, then specify the size.''. We then transform the decomposed units into natural language spans/sentences based on the paired templates. We call these spans/sentences ``procedural fragments''. 

\vspace{5pt}

\begin{figure}[t]
    \centering
    \includegraphics[width=1\linewidth]{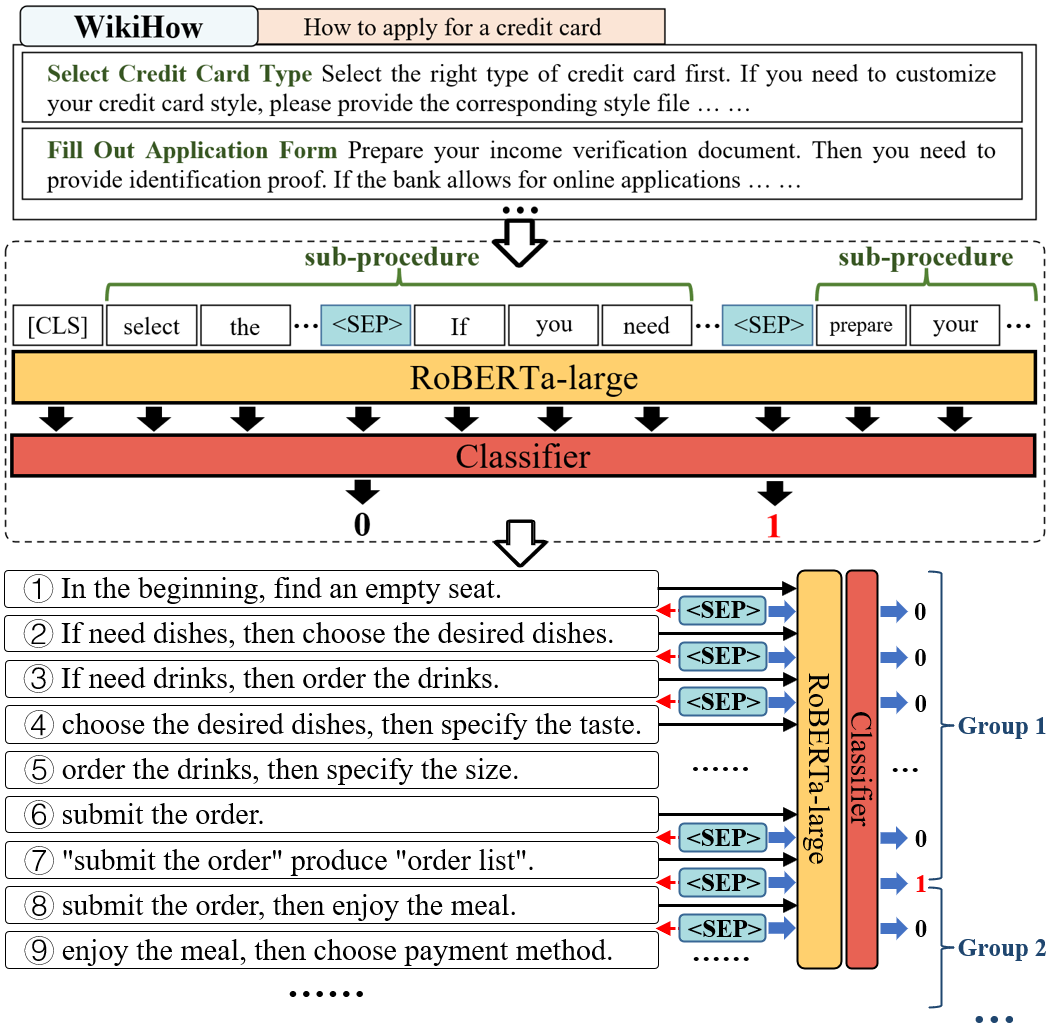}
     \caption{The grouping of procedural fragments using a pre-trained boundary identification model.}
    \label{fig:Grouping_Ordering}
\end{figure}
\vspace{5pt}

\paragraph{Grouping \& Ordering}
When writing a procedural document, it is vital to provide information in a logical way, namely, describing the whole procedure sub-procedure by sub-procedure~\cite{futrelle1999summarization, futrelle2004handling}. Hence, we need to group the fragments into sub-procedures. Specifically, we employ a simple boundary identification model, which employs the RoBERTa-large model~\cite{liu2019roberta} with a classifier layer, to predict whether a fragment is the end of a group. We train it on the WikiHow corpus~\citet{bolotova2023wikihowqa}, which consists of procedural documents collected from the \href{https://www.wikihow.com/}{WikiHow} website and format marks indicate different sub-procedures. The pre-trained model is then used to assign group marks for fragments, cf., Figure~\ref{fig:Grouping_Ordering}. 
It is worth noticing that the way people describe a procedure is not the way the machine searches over the graph. Thus, it is necessary to order the fragments to better match human expression. For example, we need to exchange fragment \ding{194} and fragment\ding{195} of Group 1 in Figure~\ref{fig:Grouping_Ordering}. We achieve this via a fragment ordering model, whose structure is borrowed from~\citet{bin2023non}. Besides the WikiHow documents, we train this model on the remaining publicly
available procedural document datasets~\cite{castelli2020techqa, zhang-etal-2020-reasoning, lyu2021goal,sakaguchi2021proscript, nandy2021question} to reorder sentences after random shuffle. We use the pre-trained model to determine the order of fragments in the same group.

\paragraph{Aggregating \& Smoothing}
In a standard procedural document, a single sentence~(``If the customer needs dishes, then choose the desired dishes and specify the taste.'') may convey multiple procedural fragments~(\ding{193} and \ding{195}). Given this, we reuse the boundary identification model to aggregate fragments that should be presented in the same sentence. The model is retrained to identify the correct ends of sentences from randomly inserted ones on the datasets used in the ordering phase. The pre-trained model is used to assign sentence marks for the fragments. 
We further add the actor information before all fragments corresponding to each actor. At this point, all fragments have been organized in proper order with group and sentence marks. We paraphrase the fragments via ChatGPT, which has shown near or even superior human-level performance in many paraphrasing task~\cite{chui2023chatgpt}. After paraphrasing, we notice that there are a small number of redundant expressions caused by repeatedly walked nodes and inconsistent actions/constraints generated by ChatGPT. Thus, we further refine the documents with a few handwriting rules and manual corrections. At last, we develop a dataset with 3,394 high-quality document-graph pairs. On average, each document contains 10.67 sentences and each sentence contains 15.22 words. See Table~\ref{dataset:statistics} in Appx.~\ref{app:Aggregating_Smoothing} for more statistics.

\subsection{Dataset Analysis}
\label{Dataset Quality Evaluation}
We analyze our datasets to investigate whether the generated procedural documents are consistent with the original graphs, whether the documents are qualified according to human standards, and whether the proposed strategies contribute to a better quality of the dataset. Therefore, we conduct both automatic and human evaluations to compare the dataset constructed by our three-stage pipeline with the datasets constructed by two variations. 
\paragraph{Automatic Evaluation} We adopt two commonly used Data2Text metrics. The \textit{FINE} score~\cite{faille2021entity} models the evaluation as a natural language inference task --- by inferring fragments from the documents it checks omissions, while the other direction checks hallucinations. The \textit{ESA} score~\cite{faille2021entity} evaluates the coverage of entities and actions in the documents. For both, higher values indicate better performance. Details are listed in the appendix~\ref{app:AutomaticEvaluation}.

\paragraph{Human Evaluation} Inspired by~\citet{miller1979humanistic}, we ask three workers to score the document from 1 to 5 on five criteria:
\textit{readability}, the quality of the document to be understood easily; 
\textit{accuracy}, the quality of the document accurately describing the information in the procedure graph; 
\textit{clarity}, the quality of the document expressing the complex execution of actions logically; 
\textit{simplicity}, the quality of the document not containing redundant information; 
\textit{usability}, the quality of the document aiding users in accomplishing this procedure.
Details are listed in the appendix~\ref{app:HumanEvaluation}. 

\paragraph{Variations} We create two variants of our three-stage pipeline: \textit{concatenating}, which directly concatenates all fragments to form the documents; \textit{paraphrasing}, which directly paraphrases the concatenation of all fragments using ChatGPT without grouping, ordering and aggregating process.  
Note that, we don't involve \textit{concatenating} in the automatic evaluation. This is because the concatenation of all fragments is always consistent with the information in the graphs but lacks fluency and logicality which requires further human evaluation. 
\begin{figure}[t]
    \centering
    \includegraphics[width=1\linewidth]{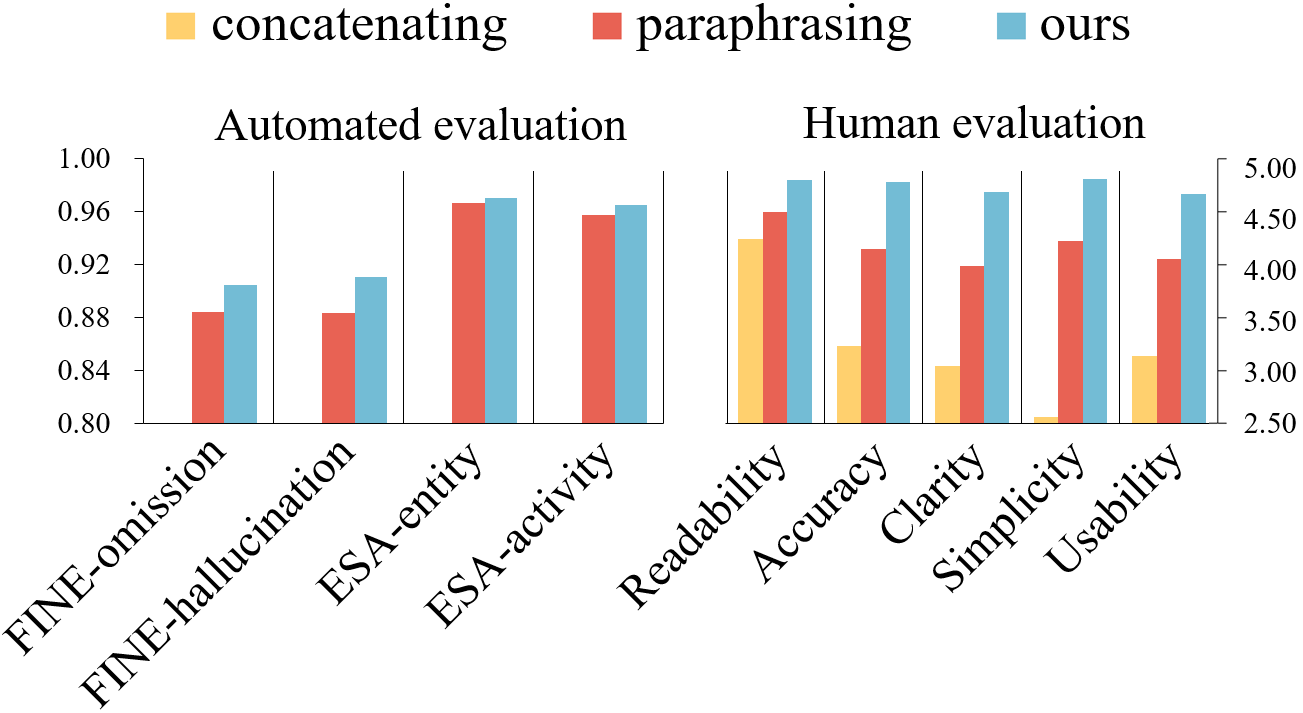}
    \caption{Comparison of our method with two variations via automatic and human evaluations.}
    \label{fig:DatasetEva&Human}
\end{figure}


\begin{table*}[t]
\caption{Performances of state-of-the-art baselines and LLMs. Higher values indicate better performances.}
\label{exp_results}
\centering
\scalebox{0.65}{
\begin{tabular}{clcccccccccc}
\hline
\multicolumn{1}{c|}{\multirow{2}{*}{\textbf{Row}}} & \multicolumn{1}{c|}{\multirow{2}{*}{\textbf{Model}}} & \multicolumn{1}{c|}{\multirow{2}{*}{\textbf{Actor}}} & \multicolumn{1}{c|}{\multirow{2}{*}{\textbf{Action}}} & \multicolumn{2}{c|}{\textbf{Constraint}}              & \multicolumn{3}{c|}{\textbf{Gateway}}                                            & \multicolumn{3}{c}{\textbf{Flow}}                            \\ \cline{5-12} 
\multicolumn{1}{c|}{}                              & \multicolumn{1}{c|}{}                                & \multicolumn{1}{c|}{}                                & \multicolumn{1}{c|}{}                                 & \textbf{Data}  & \multicolumn{1}{c|}{\textbf{Action}} & \textbf{Exclusive} & \textbf{Inclusive} & \multicolumn{1}{c|}{\textbf{Parallel}} & \textbf{Sequence} & \textbf{Condition} & \textbf{Constraint} \\ \hline
\multicolumn{1}{c|}{1}                             & \multicolumn{1}{l|}{\citet{sonbol2023machine}}               & \multicolumn{1}{c|}{0.028}                           & \multicolumn{1}{c|}{0.308}                            & 0.213          & \multicolumn{1}{c|}{-}               & 0.485              & -                  & \multicolumn{1}{c|}{0.279}             & 0.056             & 0.047              & 0.017               \\ \cline{2-12} 
\multicolumn{1}{c|}{2}                             & \multicolumn{1}{l|}{\citet{neuberger2023beyond}}             & \multicolumn{1}{c|}{0.027}                           & \multicolumn{1}{c|}{0.276}                            & -              & \multicolumn{1}{c|}{-}               & 0.469              & -                  & \multicolumn{1}{c|}{0.337}             & 0.074             & 0.061              & -                   \\ \cline{2-12} 
\multicolumn{1}{c|}{3}                             & \multicolumn{1}{l|}{\citet{sholiq2022generating}}            & \multicolumn{1}{c|}{-}                               & \multicolumn{1}{c|}{0.387}                            & -              & \multicolumn{1}{c|}{-}               & 0.463              & -                  & \multicolumn{1}{c|}{0.198}             & 0.091             & 0.022              & -                   \\ \cline{2-12} 
\multicolumn{1}{c|}{4}                             & \multicolumn{1}{l|}{PET~\cite{bellan2023pet}}                             & \multicolumn{1}{c|}{0.085}                           & \multicolumn{1}{c|}{0.430}                            & 0.069          & \multicolumn{1}{c|}{-}               & \underline{0.493}              & -                  & \multicolumn{1}{c|}{-}                 & 0.164             & 0.026              & 0.000                    \\ \cline{2-12} 
\multicolumn{1}{c|}{5}                            & \multicolumn{1}{l|}{CIS~\cite{bellan2022leveraging}}                             & \multicolumn{1}{c|}{0.633}                           & \multicolumn{1}{c|}{0.639}                            & -              & \multicolumn{1}{c|}{-}               & 0.455              & -                  & \multicolumn{1}{c|}{-}                 & 0.203             & 0.157              & -                   \\ \hline
\multicolumn{1}{l|}{}                              & \multicolumn{1}{l|}{Flan-T5~\cite{chung2022scaling}}                         & \multicolumn{1}{c|}{}                                & \multicolumn{1}{c|}{}                                 &                & \multicolumn{1}{c|}{}                &                    &                    & \multicolumn{1}{c|}{}                  &                   &                    &                     \\
\multicolumn{1}{c|}{6}                             & \multicolumn{1}{l|}{\quad + Few-shot In-context Learning}               & \multicolumn{1}{c|}{0.206}                           & \multicolumn{1}{c|}{0.362}                            & -              & \multicolumn{1}{c|}{-}               & 0.376              & -                  & \multicolumn{1}{c|}{-}                 & 0.084             & 0.013              & -                   \\
\multicolumn{1}{c|}{7}                             & \multicolumn{1}{l|}{\quad + Supervised Fine-tuning}                     & \multicolumn{1}{c|}{\underline{0.659}}                           & \multicolumn{1}{c|}{\underline{0.684}}                            & 0.589          & \multicolumn{1}{c|}{\underline{0.366}}           & 0.419              & 0.045              & \multicolumn{1}{c|}{\underline{0.393}}             & 0.395             & \underline{0.168}              & 0.363               \\ \hline
\multicolumn{1}{l|}{}                             & \multicolumn{1}{l|}{ChatGPT~\cite{ouyang2022training}}                         &  \multicolumn{1}{c|}{}                                 & \multicolumn{1}{c|}{}                                 &                & \multicolumn{1}{c|}{}                &                    &                    & \multicolumn{1}{c|}{}                  &                   &                    &                     \\

\multicolumn{1}{c|}{8}                             & \multicolumn{1}{l|}{\quad + Few-shot In-context Learning} &\multicolumn{1}{c|}{0.625}                           & \multicolumn{1}{c|}{0.681}                            & \underline{0.687}          & \multicolumn{1}{c|}{0.286}           & 0.477              & \textbf{0.173}     & \multicolumn{1}{c|}{0.388}             & \underline{0.408}             & 0.158              & 0.\underline{444}               \\ \hline
\multicolumn{1}{l|}{}                              & \multicolumn{1}{l|}{Llama2~\cite{touvron2023llama}}                          & \multicolumn{1}{c|}{}                                & \multicolumn{1}{c|}{}                                 &                & \multicolumn{1}{c|}{}                &                    &                    & \multicolumn{1}{c|}{}                  &                   &                    &                     \\
\multicolumn{1}{c|}{9}                             & \multicolumn{1}{l|}{\quad + Few-shot In-context Learning}               & \multicolumn{1}{c|}{0.502}                           & \multicolumn{1}{c|}{0.573}                            & 0.357          & \multicolumn{1}{c|}{0.049}           & 0.446              & 0.067              & \multicolumn{1}{c|}{0.128}             & 0.193             & 0.107              & 0.201               \\
\multicolumn{1}{c|}{10}                            & \multicolumn{1}{l|}{\quad + Supervised Fine-tuning}                     & \multicolumn{1}{c|}{\textbf{0.674}}                  & \multicolumn{1}{c|}{\textbf{0.744}}                   & \textbf{0.779} & \multicolumn{1}{c|}{\textbf{0.499}}  & \textbf{0.554}     & \underline{0.090}              & \multicolumn{1}{c|}{\textbf{0.398}}    & \textbf{0.478}    & \textbf{0.319}     & \textbf{0.467}      \\ \hline
\multicolumn{12}{l}{\small * The best results are marked in bold, and the second-best results are marked with underlines.}                                                                                                                                                                                                                                                                                                                 
\end{tabular}
}
\vspace{5pt}
\end{table*}

\paragraph{Results}

As shown in Figure~\ref{fig:DatasetEva&Human}, the proposed three-stage pipeline achieves better performance than the other two variations under both automatic and human evaluation. Automatic evaluation reveals that although the paraphrasing variation obtains almost close ESA scores to our pipeline, it loses the game completely on the FINE metrics without the help of grouping, ordering and aggregating strategies. From human evaluation, we further demonstrate the superiority of our pipeline to describe procedural graphs in a fluent, accurate, logical, simple, and user-friendly way. Specifically, the concatenating variation gets the lowest scores on all criteria due to its inaction of unreadable spans, redundant information, and chaotic orders. In line with the automatic evaluation, we observe the largest gap between the paraphrasing variation and our pipeline on the clarity criterion due to its invisibility of complex logic among the fragments. Note that, the ICC scores of all evaluations are above 0.75, indicating the reliability of human results~\cite{koo2016guideline}.

\section{Experiments}
We conduct systematic experiments to answer \textit{Q1} and \textit{Q2} raised in Sec.~\ref{sec:intro}. We split the dataset into train, validation and test sets with 3:1:2 ratio. For \textit{Q1}, we collect state-of-the-art baselines and evaluate them in \benchmark. We further introduce three metrics based on the underlying structure of graphs and the surface form of elements. Specifically, we use the BLEU scores to measure the performance of actor, action and constraint extraction and F1 scores to measure the performance of gateway prediction. Besides, the performance of flow prediction is measured via soft F1 scores~\cite{tandon-etal-2020-dataset}, which is computed based on the BLEU scores of associated textual elements. For \textit{Q2}, we involve Flan-T5~\cite{chung2022scaling}, ChatGPT~\cite{ouyang2022training} and Llama2~\cite{touvron2023llama} in \benchmark to show their potentials and improve their performance using a self-refine strategy.

\subsection{Performance of Baselines~(\textit{Q1})}

\paragraph{Baselines}
We collect five baselines:
\begin{itemize}
    \item  \citet{sonbol2023machine} uses rules to extract sequential, exclusive, and parallel actions with a few data constraints.
    \item \citet{neuberger2023beyond} designs a pipeline to extract sequential actions and organize partial non-sequential actions, ignoring all constraints and inclusive gateways.
    \item \citet{sholiq2022generating} extracts actions, exclusive gateways, and parallel gateways based on a pre-defined representation of the procedural graph.
    \item  PET~\cite{bellan2023pet} trains a sequence tagging model to extract actions with a few constraints and uses rules to construct the final graph.
   \item  CIS~\cite{bellan2022leveraging} presents a rough attempt to use LLMs\footnote{The LLM used in \citet{bellan2022leveraging} is GPT3. We replace it with ChatGPT in the current setting for a fair comparison.} for action extraction via few-shot in-context learning and constructs the graphs via  handwritten rules. 
\end{itemize}

\paragraph{Results}
As shown in Table~\ref{exp_results}~(Rows 1-5), existing studies are far from solving this task well, especially when organizing the logical structure of graphs~(cf., the results on gateways and flows). This is because either rules or neural models are derived from limited data, leading to an incomplete coverage of all elements and an incomprehensive understanding of complex documents. Additionally, we have the following observations: \par
1) Heuristic methods~(Row 1-3) perform poorly for actor, action and constraint extraction. The reason is that hand-written rules fail to understand various expressions and coreferences, resulting in poor generalization. \par
2) PET~(Row 4), though being a customized deep neural model, only performs slightly better than heuristic methods. This is because the PET model is only trained on 45 samples and utilizes several rules to construct the flows. \par
3)  We conjecture the reason why all baselines only meet part of the requirements of optimal procedural graphs lies in the huge cost of writing rules and annotating data. We believe this issue can be alleviated with the dataset created in Sec.~\ref{sec:data}.\par
4)  It is not surprising that CIS~(Row 5) achieves the highest scores on the action and actor extraction among these baselines, as it gains more power to understand word meanings from the LLM. Note that, the increase in flow prediction between PET and CIS also comes from more accurate action extraction instead of more delicate rules to construct the graph. This increase further encourages us to investigate more potentials of LLMs in procedural graph extraction. \par
5)  All baselines perform poorly on flow prediction, indicating the challenge of understanding logical structures in documents. This motivates us to find out, besides an in-depth study of LLMs, what else matters to construct the logic in graphs. 

\subsection{Performance of LLMs~(\textit{Q2})}
\paragraph{LLMs}
We investigate three advanced LLMs. The first one is ChatGPT~\cite{ouyang2022training}, which is good at information comprehension and text generation. Different from CIS~\cite{bellan2022leveraging}, we use our high-quality dataset to apply few-shot in-context learning~(ICL) on ChatGPT. The other two~(Flan-T5~\cite{chung2022scaling} and Llama2~\cite{touvron2023llama}) are open-source alternatives to ChatGPT. Besides ICL, we also deploy the alternatives with supervised fine-tuning strategies.

\paragraph{Results}
As expected, the LLMs~(Rows 6-10) update state-of-the-art results on all metrics, especially for actor, action and constraint extraction. However, for gateway and flow predictions, all LLMs can hardly get $>0.5$ F1 scores, exhibiting their weakness in arranging logical structures of graphs. We also have the following observations: \par
1) Our high-quality dataset significantly promotes LLMs' ability on procedural graph extraction. A piece of direct evidence is that we see a rising trend between the results on Rows 5 and 8, whose only difference lies in the data used in the few-shot ICL. Another piece of evidence is that the gap between Flan-T5 and Llama2 is rapidly narrowed after using more data from our dataset for tuning. \par
2) The procedural knowledge, especially the logical structure of non-sequential actions, has been overlooked during the initial training of LLMs. This is demonstrated by the fact that fine-tuned LLMs perform better than few-shot LLMs. The Llama2 model even beats ChatGPT after learning more procedural knowledge from supervised fine-tuning. \par
3) LLMs, including CIS, show significant potential for actor, action and constraint extraction, indicated by the large improvements compared with Rows 1-4. This is because LLMs are good at understanding lengthy contexts and thus have the advantage of identifying meaningful elements from procedural documents. We also believe that the emergence of more powerful LLMs in the future will continue to promote better results on these metrics. \par
4) We believe the biggest challenge for LLMs to extract accurate procedural graphs lies in the lack of logic reasoning ability among actions, especially non-sequentially executed ones. The supporting evidence is that, though largely surpassing baselines on the actor, action and constraint extractions, LLMs don't present such impressive improvements in gateway extraction. This implies that we can boost LLMs' performances by paying extra attention to non-sequential logic prediction.

\begin{figure}[!t]
    \centering
    \includegraphics[width=1\linewidth]{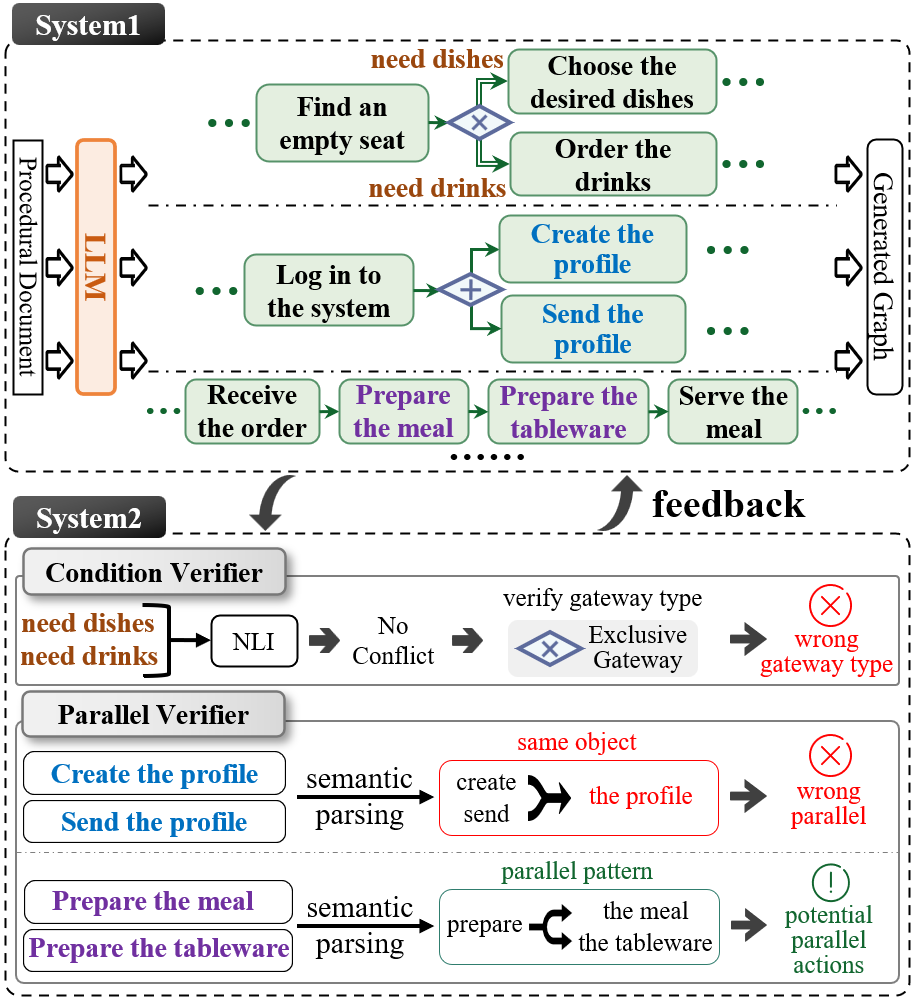}
    \caption{The self-refine strategy, in which ``System1'' extracts procedural graphs and ``System2'' verifies gateways of graphs and provides feedback for refinement.
    }
    \label{fig:method}
\end{figure}

\subsection{Self-refine Strategy}
To overcome the above-mentioned challenge, inspired by \citet{nye2021improving, madaan2023self}, we design a self-refine strategy to help LLMs gain logic reasoning ability among actions from iterative feedback and refinement. As shown in Figure~\ref{fig:method}, it consists of two systems --- ``System1'' is used to directly extract procedural graphs from documents, ``System2'' is used to verify the extracted graphs and provide feedback for further refinement of ``System1''. In ``System2'', we center on the shortest slab and carefully examine the gateway prediction results using the condition and parallel verifiers.

\paragraph{Condition Verifier} We design the condition verifier to handle both exclusive and inclusive gateways, whose key difference lies in the conditions followed by gateways. It is worth noticing that, with the exclusive gateway, there is always one and only one condition that can be met. Accordingly, we suppose if the conditions hold conflict, the gateway should be the exclusive one; otherwise, it should be the inclusive one. Particularly, we use a pre-trained natural language inference~(NLI) model~\cite{liu2019roberta} to detect the conflict and verify gateways. For example, after feeding ``need dishes'' and ``need drinks'' into the NLI model, we get ``No Conflict''. This suggests the gateway should be the inclusive one, which is different from the result predicted by ``System1''. In this case, the condition verifier is triggered to provide feedback to refine ``System1''.

\paragraph{Parallel Verifier} We further design the parallel verifier to reorganize the actions performed in parallel. We notice that if the actions act on the same object, they can never be executed in parallel. Given that, we extract the objects of actions using the semantic parsing tool and determine the parallel gateway based on these objects. For example, as ``create the profile'' and ``send the profile'' have the same object ``the profile'', they cannot be performed in parallel. Besides, the optimal procedural graph is expected to help users in a time-saving way~\cite{miller1979humanistic}. Thus, we further prepare another type of feedback by examining the sequential actions. For example, as ``prepare the meal'' and ``prepare the tableware'' have different objects, they have a big chance to be performed in parallel. We provide ``System1'' with both types of feedback for the refinement of mistakes on parallel gateways. 

\begin{figure}[t]
    \centering
    \includegraphics[height=3.6cm]{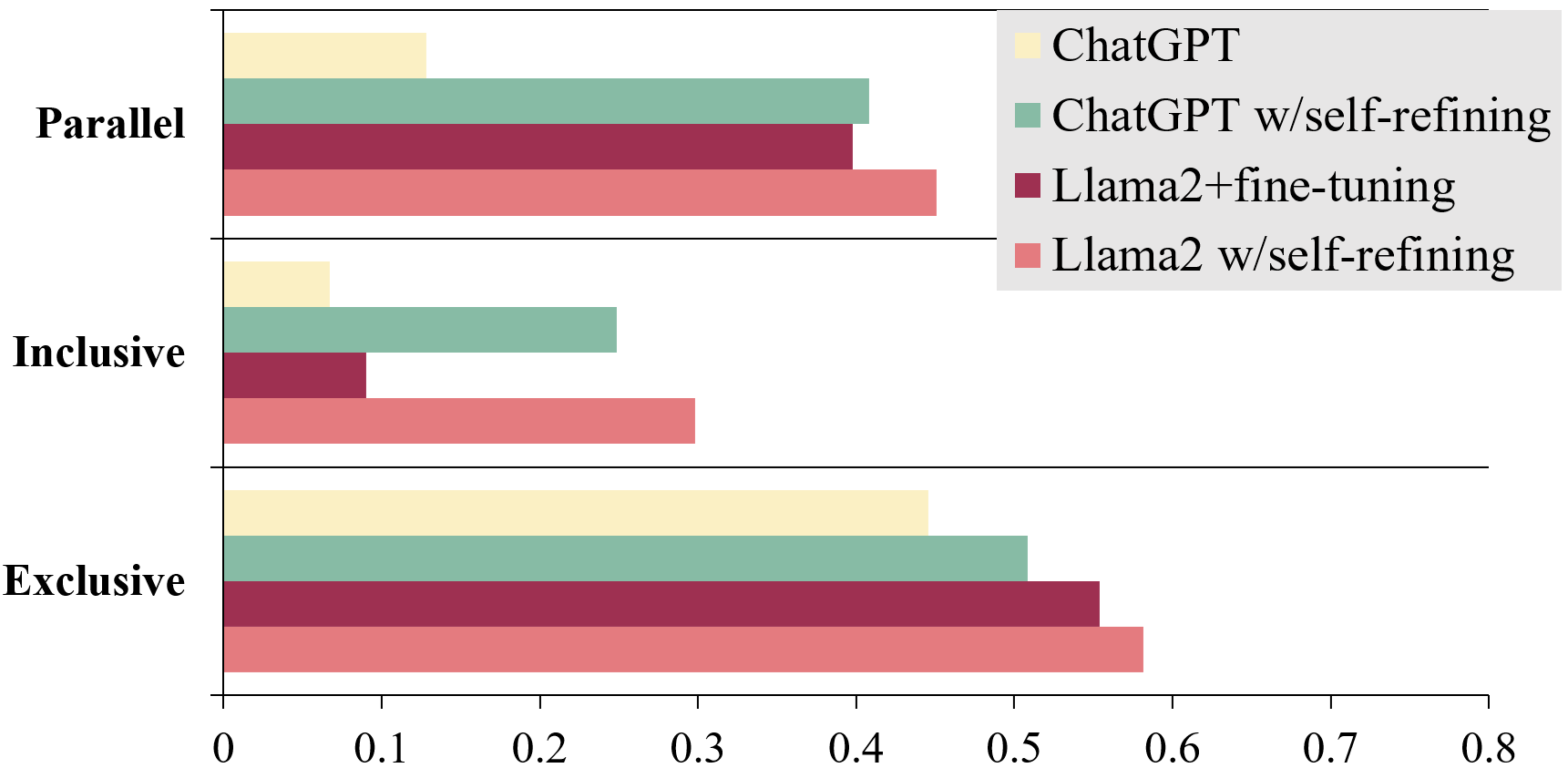}
    \caption{F1 scores on gateway predictions of LLMs and their variants with our self-refine strategy.
    }
    \label{fig:improve}
\end{figure}

\paragraph{Results}
We apply the self-refine strategy to the top-two winners in Table~\ref{exp_results}, i.e., ChatGPT and fine-tuned Llama2. As shown in Figure~\ref{fig:improve}, both ChatGPT and fine-tuned Llama2 have better performances with the help of our self-refine strategy. More surprisingly, there are significant improvements in inclusive gateway extraction, whose previous scores are extremely poor. This indicates that, with effective strategies, LLMs have the potential to gain logic reasoning ability among actions including non-sequential ones. The improvement of Llama2 is not as large as that of ChatGPT. This is because the model's gain from the self-refine strategy drops as it encounters more procedural knowledge. This suggests the importance of incorporating the learning of procedural knowledge during the pre-training stage of LLMs, which will be beneficial for LLMs' logic reasoning ability.

\begin{figure}[t]
    \centering
    \includegraphics[height=4cm]{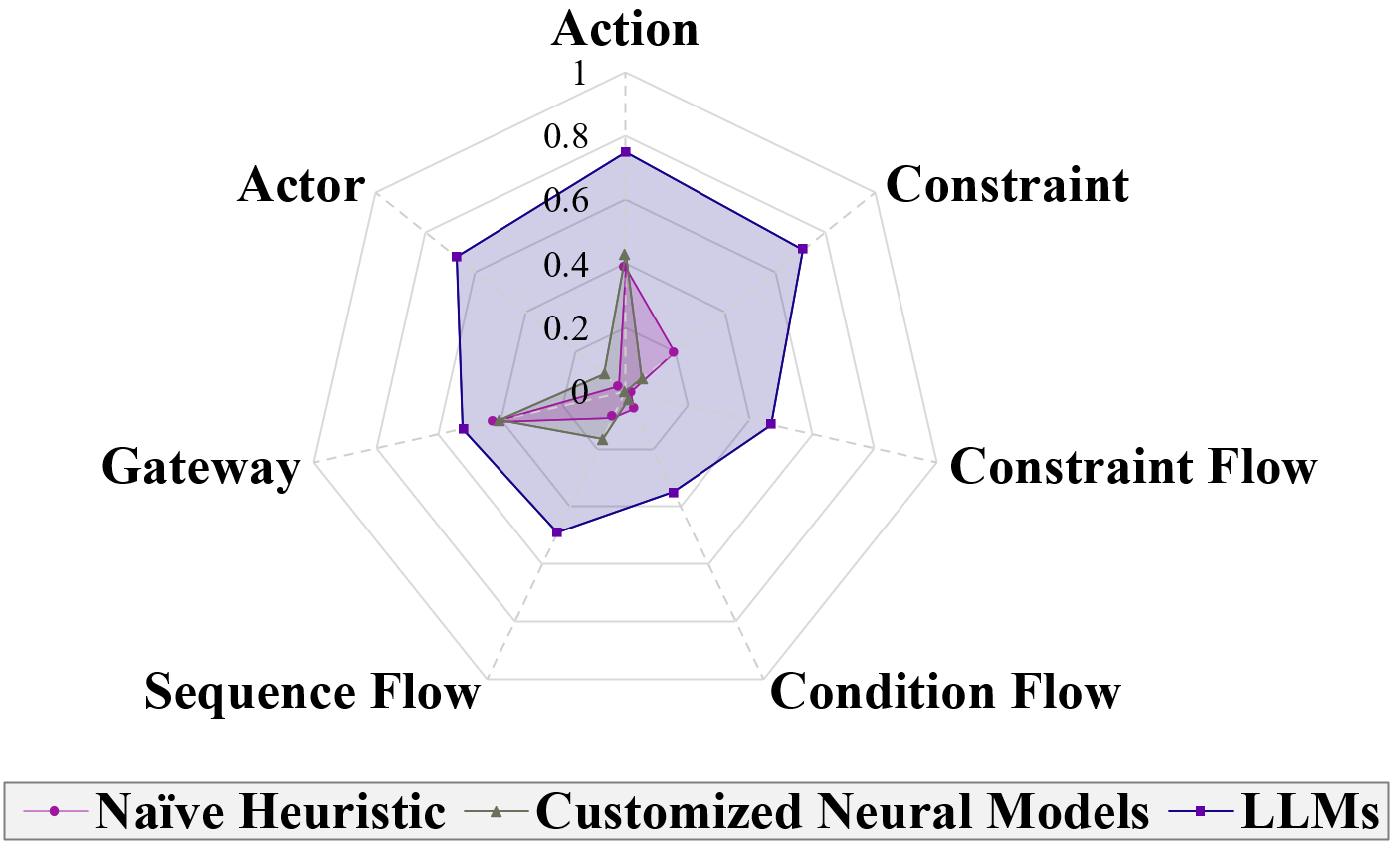}
    \caption{Best performances of heuristics, customized neural models and LLMs on seven dimensions.
    }
    \label{fig:Radar}
\end{figure}

\paragraph{Auxiliary Analysis}
We group the methods into three sets, i.e., the na\"ive heuristics~(rows 1-3), customized neural models~(row 4), and LLMs~(rows 5-10). Towards a clear understanding of the advantages and challenges, we report the best performances of each type of method on seven dimensions in Figure~\ref{fig:Radar}. Even with the highest scores, heuristic methods and customized neural models exhibit poor performance across all dimensions and can hardly handle condition and constraint flows due to their neglect of the logical structure in documents. LLMs show substantial improvements compared to the others in all dimensions except for the gateway. This suggests that even the powerful LLMs face challenges in managing non-sequential actions, which is also the main challenge when conducting optimal procedural graph extraction.

\section{Conclusion and Future work}
We propose the \benchmark benchmark, where we systematically study the progress of current procedural graph extraction methods and explore the potential of emerging LLMs on this task. We equip \benchmark with a high-quality dataset, which is about ten times larger than the previous largest ones. Experimental results of baselines in \benchmark reveal that current methods are far from solving this task well. We further involve three advanced LLMs in \benchmark to demonstrate their advantages in extracting textual elements and challenges in organizing local structures. To overcome the main challenge, we design a novel self-refine strategy to empower the LLMs' ability in reasoning gateways. The results show that, with effective strategies, LLMs have the potential of LLMs to comprehend the logical structure among non-sequential actions.


We hope \benchmark can benefit the research on optimal procedural graphs extraction. There are several directions for further work. 
First, to improve the performance of LLMs, we suggest introducing procedural knowledge during the pre-training stage of LLMs. 
Second, to gain better flow extraction, we will explore more effective methods for handling complex logic structures in the documents. 
Lastly, we plan to find a real-world scenario to investigate, besides accuracy, what else is limiting the practical usage of automatic procedure graph extraction.

\section{Ethics Statement}
The dataset we constructed is sourced from a publicly available model collection originated from the BPM Academic Initiative and does not contain any sensitive or personal privacy related information. 
The procedural graphs used in our dataset are available for research purposes on the ``CC BY-NC-SA 3.0 DEED'' licence~\footnote{\url{http://fundamentals-of-bpm.org/process-model-collections/}}, which explicitly permits that we can not only use the collected examples but also ``remix, transform, and build upon the material''. 
Therefore, we believe that there is no ethical issue with our work. 
\section{Limitations}
The procedural documents in our dataset can hardly be equal to the documents directly written by experts. Despite this fact, we believe our dataset can still largely promote the study of automatic procedural graph extraction. The samples for few-shot learning are randomly selected from the dataset. This may lead to fluctuations in LLMs' results. However, since the samples of all LLMs are randomly chosen, we believe the experimental setup is fair. We acknowledge that carefully selecting samples would yield better results, but this is not the focus of this benchmark. We only design one prompt for all LLMs. Although using another elaborate prompt could introduce new variations to the experimental results, we consider this as a topic for future research.

\section*{Acknowledgement}
This work was supported in part by the National Natural Science Foundation of China (No. 62206191 and No. 62272330);
in part by the Natural Science Foundation of Sichuan (No. 2023NSFSC0473), 
and in part by the Fundamental Research Funds for the Central Universities (No. 2023SCU12089 and  No. YJ202219).


\bibliography{anthology}

\clearpage
\appendix



\section{Details of Data Construction}
\label{app:DataConstruction}

\begin{table*}[t]
\caption{Designed templates used to transfer the decomposed units into natural language procedural fragments. 
Due to the fact that the gateways in the procedural graphs are paired, each pair of gateways includes a ``branch gateway'' representing the beginning of the non-sequential execution of actions and a ``merge gateway'' representing the end of the non-sequential execution. We use prefix ``B\_'' to indicate the branch gateway, and prefix ``M\_'' to indicate the merge gateway. 
In addition, the ``Flow'' in a unit is used to connect different elements, while the ``condition'' in a unit represents that this Flow connects exclusive or inclusive gateways with other elements, so there exists specific condition on this Flow. 
For simplicity, we use ``XOR'', ``OR'' and ``AND'' as abbreviations for exclusive, inclusive and parallel gateways respectively. 
}
\label{tab:triple_templates}

\centering
\scalebox{0.95}{
\begin{tabular}{|l|l|}
\hline
\multicolumn{1}{|c|}{\textbf{Decomposed Unit}}             & \multicolumn{1}{c|}{\textbf{Template}}                      \\ \hline
(Start, Flow, Action)                                  & In the beginning, \{Action\}.                               \\ \hline
(Start, Flow, B\_Gateway)                              & In the beginning,                                           \\ \hline
(Action1, Flow, Action2)                               & \{Action1\}, then \{Action2\}.                                                \\ \hline
(Action, Flow, End)                                    & \{Action\}, and the procedure ends.                         \\ \hline
(B\_XOR, condition, Action)                            & If \{condition\}, then \{Action\}.                          \\ \hline
(B\_XOR, condition, B\_Gateway)                        & If \{condition\},                                           \\ \hline
(B\_XOR, condition, M\_Gateway, Flow,   Action)        & If \{condition\}, then \{Action\}.                          \\ \hline
(B\_XOR, condition, End)                               & If \{condition\}, then the procedure ends.                  \\ \hline
(B\_OR, condition, Action)                             & If \{condition\}, then \{Action\}.                          \\ \hline
(B\_OR, condition, B\_Gateway)                         & If \{condition\},                                           \\ \hline
(B\_OR, condition, M\_Gateway, Flow,   Action)         & If \{condition\}, then \{Action\}.                          \\ \hline
(B\_OR, condition, End)                                & If \{condition\}, then the procedure ends.                  \\ \hline
\makecell[l]{(B\_AND, Flow, $*^1$)\\(B\_AND, Flow, $*^2$)}                   & \{$*^1$\}, at the same time, \{$*^2$\}.                           \\ \hline
\makecell[l]{(B\_AND, Flow, $*^1$)\\(B\_AND, Flow, $*^2$)\\(B\_AND, Flow, $*^3$)} & \{$*^1$\}, at the same time, \{$*^2$\}, meanwhile, \{$*^3$\}.        \\ \hline
(M\_Gateway, Flow, Action)                             & \{Action\}.   
\\ \hline
(M\_Gateway, Flow, End)                                & The procedure ends.                                         \\ \hline
(Action, Flow, DataConstraint)                         & ``\{Action\}'' produce ``\{DataConstraint\}''.              \\ \hline
(DataConstraint, Flow, Action)                         & ``\{Action\}'' require access to ``\{DataConstraint\}''.    \\ \hline
(Action, Flow, ActionConstraint)                       & For \{Action\}, pay attention to that \{ActionConstraint\}. \\ \hline
\end{tabular}
}
\vspace{10pt}
\end{table*}

\paragraph{Decomposition \& Transformation}
\label{app:DT}
To narrow the huge gap between the complex graph and the length document and meanwhile maintain the consistency between the transferred fragments and the original graphs, such as the texts of the entities, actions, etc., we design hand-written templates to transfer the graphs into natural language fragments. The designed templates are listed in Table~\ref{tab:triple_templates}. 
Additionally, we find that not all units on the original graphs are meaningful and need to be explicitly expressed in generated procedural documents. So we filter out those meaningless units through heuristic rules.

\paragraph{Aggregating \& Smoothing}
\label{app:Aggregating_Smoothing}

\begin{figure*}[th]
    \centering
    \includegraphics[width=\textwidth]{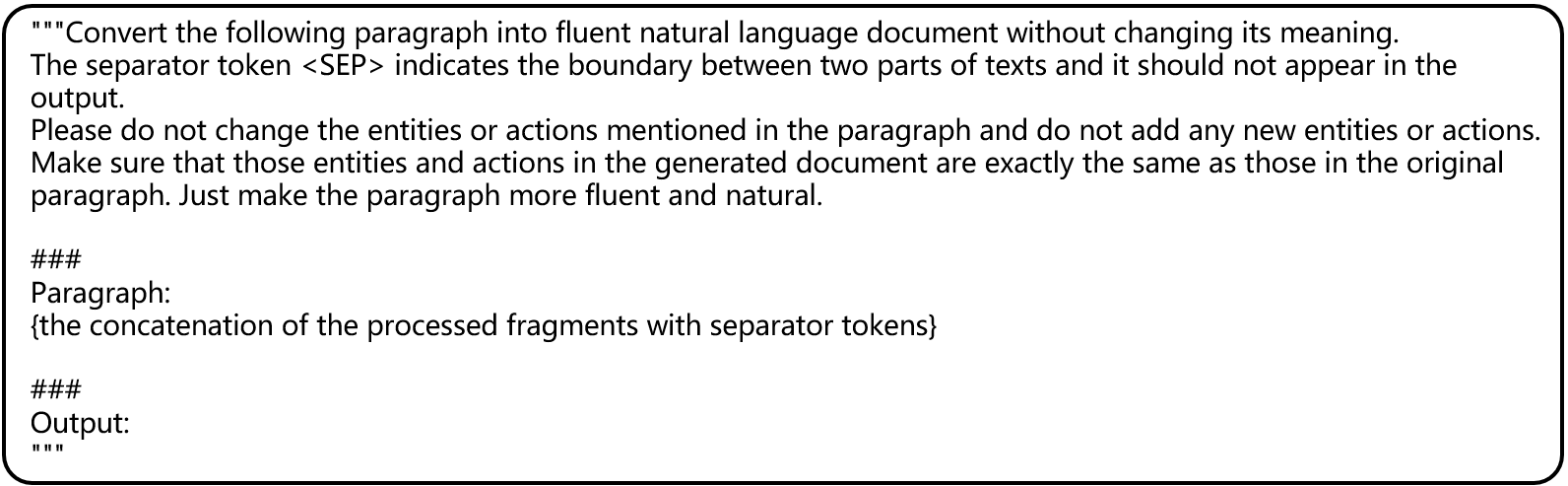}
    \caption{The designed prompt used to rephrase the concatenation of the processed fragments. 
    }
    \label{fig:Rephrase}
    \vspace*{-0cm}
\end{figure*}

We conduct rephrasing operation on the concatenation of the processed fragments using LLM. We add the unique separator token ``<SEP>'' to indicate the group and sentence marks of the fragments, which can prompt the model to describe these fragments logically and coherently. 
Moreover, we add the actor information before all fragments corresponding to each actor. For example, we add the text ``For the customer:'' before all fragments corresponding to the customer. 
We adopt ChatGPT~(gpt-3.5-turbo) to conduct the rephrasing operation. The designed prompt is shown in Figure~\ref{fig:Rephrase}. 
At last, we develop a dataset with 3,394 high-quality document-graph pairs. 
Additionally, although it is difficult to exactly define the large of the dataset, we argue our dataset is large enough because it is about ten times larger than the previous largest datasets and successfully supports the evaluation of current studies and the discovery of future directions in this field. 
We present the statistics information of the constructed dataset in Table~\ref{dataset:statistics}.

\section{Details of Dataset Quality Evaluation}
\label{app:DatasetQualityEvaluation}

\paragraph{Automatic Evaluation}
\label{app:AutomaticEvaluation}
Following the evaluation strategies commonly used in Data2Text task~\cite{lin2023survey}, we conduct automatic evaluation to evaluate whether the generated procedural documents provide information consistent with the original graphs. The automatic metrics used for evaluation are listed as follows: 

\textbf{FINE}~\cite{duvsek2020evaluating}: evaluating the semantic equivalence of generated documents with a natural language inference model~\footnote{\url{https://huggingface.co/FacebookAI/roberta-large-mnli}}. The natural language inference model can be used to determine whether a ``hypothesis'' is true~(entailment) given a ``premise''. 
We first compute the entailment between the generated document and the transferred fragments to evaluate whether the generated document fully covers the original graph's information~(omission). We take the generated document as the ``premise'' and use the natural language inference model to determine whether the transferred fragments are true~(entailment), i.e., whether the semantics of the transferred fragments are covered by the generated document. 
Then we exchange their positions to evaluate whether the generated document not contains redundant information beyond the graph~(hallucination). 

\textbf{ESA}~\cite{faille2021entity}: evaluating the faithfulness of the generated documents based on the coverage of the entities and actions in the original graphs. We use named entity recognition tool~\footnote{\url{https://huggingface.co/dslim/bert-base-NER}} to extract the entities in the original graphs. Then we evaluate whether the entities and actions in the original graph are covered by the generated document with exact lexical match. 


\paragraph{Human Evaluation}
\label{app:HumanEvaluation}
We design five criteria to evaluate the generated documents through human scoring. 
To ensure the reliability of human evaluation, we hire three domain experts to evaluate the generated documents and calculate the ICC~(Intraclass Correlation Coefficient)~\cite{shrout1979intraclass} score between different experts. The higher ICC score indicates higher consistency between different experts and higher reliability of the evaluation. Generally an ICC of $0.75$ or higher indicates that the evaluation is reliable~\cite{koo2016guideline}. ICC is calculated as follows: 

\begin{equation}
    \text{ICC} = \frac{\text{MS}_{\text{between}} - \text{MS}_{\text{within}}}{\text{MS}_{\text{between}} + (k-1) \times \text{MS}_{\text{within}}}
\end{equation}

\noindent
where $\text{MS}_{\text{between}}$ is the mean square for between groups variability, $\text{MS}_{\text{within}}$ is the mean square for within groups variability and $k$ is the number of groups. And we ensure that the ICC scores for all of our evaluations are $\geq 0.75$. 



\begin{table}[!th]
\caption{Statistics of the constructed dataset. We present statistical information on the number of documents and various elements in the dataset.}
\label{dataset:statistics}
\centering
\scalebox{0.79}{
\begin{tabular}{|c|c|cc}
\hline
\textbf{Statistics} & \textbf{Num} & \multicolumn{1}{c|}{\textbf{Statistics}} & \multicolumn{1}{c|}{\textbf{Num}} \\ \hline
Document            & 3394         & \multicolumn{1}{c|}{Data Constraint}     & \multicolumn{1}{c|}{3500}         \\ \hline
Sentence            & 37226        & \multicolumn{1}{c|}{Action Constraint}   & \multicolumn{1}{c|}{2307}         \\ \hline
Token               & 566639       & \multicolumn{1}{c|}{Sequence Flow}       & \multicolumn{1}{c|}{36438}        \\ \hline
Action              & 36537        & \multicolumn{1}{c|}{Condition Flow}      & \multicolumn{1}{c|}{10598}        \\ \hline
Exclusive Gateway   & 7024         & \multicolumn{1}{c|}{Constraint Flow}     & \multicolumn{1}{c|}{5807}         \\ \hline
Inclusive Gateway   & 1204         & \multicolumn{1}{c|}{Actor}               & \multicolumn{1}{c|}{22775}        \\ \hline
Parallel Gateway    & 2050         & \multicolumn{1}{l}{}                     & \multicolumn{1}{l}{}              \\ \cline{1-2}
\end{tabular}
}
\end{table}

\section{Details of Experiments}
\label{app:Experiments}

\subsection{Model Details}
\label{app:ModelDetails}
Heuristic models do not require training data, so we conduct evaluation for heuristic models according to the original papers' setting. The implementation details of other models are listed as follows:

\begin{enumerate}
    \item[--] \textbf{PET}: We use Roberta-large~\cite{liu2019roberta} as the backbone-model and train the model on the publicly available dataset~\footnote{\url{https://huggingface.co/datasets/patriziobellan/PET}}. A total of 3 epochs are trained, the adopted optimizer is AdamW and the learning rate is set to 5e-6. 
    
    \item[--] \textbf{CIS}: We use ChatGPT~\cite{ouyang2022training} as the pre-trained language representation model for in-context learning. We use the best-performed prompt templates following the original paper to conduct extraction with the model. 
    
    \item[--] \textbf{Flan-T5}: We fine-tune the Flan-t5-xxl model on our train set data with low-rank adaptation strategy Lora~\cite{hu2021lora}. A total of 10 epochs are trained and the learning rate is set to 1e-4. 
    
    \item[--] \textbf{Llama2}: Similar to Flan-T5 model, we fine-tune the Llama-2-70b-chat-hf model on our train set data with Lora. A total of 10 epochs are trained and the learning rate is set to 2e-4. 

    \item[--] \textbf{ChatGPT}: We use the gpt-3.5-turbo model to conduct conversation through the API provided by OpenAI~\footnote{\url{https://platform.openai.com/docs/api-reference/chat/create}}. 
    We set temperature to zero and fix seed as 42 to eliminate the influence of random sampling of the model and enable stable reproduction. 
\end{enumerate}

\subsection{Implementation Details}
\label{app:Implementation}
We split the dataset into train, validation and test sets with 3:1:2 ratio and evaluate the performance of all models uniformly on the test set. Heuristic models do not require training data, and PET trains the model on the sequence tagging data provided by itself. Therefore, we directly conduct evaluation for these models on our test set according to the original papers' setting. End to End models utilize our train set data to fine-tune the model or construct examples for few-shot in-context learning, and use the validation set for model selection~\cite{raschka2018model}. 
All the training process are conducted on a machine with $8$ $\times$ NVIDIA RTX A6000 GPUs. We use Copilot as an aid for coding. 

To facilitate the LLMs to extract the procedural graphs with text generation, 
we use a dot language~\cite{gansner2006drawing} based graph representation in the form of ``Element -> (condition) Element'' to represent the extracted graphs. We use ``XOR'', ``OR'' and ``AND'' as abbreviations for exclusive, inclusive and parallel gateways respectively, and use numbers as suffixes to distinguish different gateways of the same type on the graph~(e.g., ``OR1 -> (need dishes) Choose the desired dishes''). 
Additionally, we require the model to output the actors of corresponding extracted actions for actors predictions. 
Moreover, to make good use of the capabilities of LLMs, we adopt the few-shot in-context learning~(ICL) and Chain-of-thought~(CoT)~\cite{wei2022chain} strategy to guide the reasoning process of extracting procedural graphs from documents for the LLMs, especially for the organization of non-sequential actions. The adopted prompt consisting of elaborate instruction and three examples is shown in Figure~\ref{fig:GenerationPrompt}.



\subsection{Metrics Details}
We introduce three metrics based on the underlying structure of graphs and the surface form of elements to evaluate the model's extraction of procedural graphs. 

For actor, action and constraint extraction, we adopt the F1 based BLEU score~\cite{liang2023knowing} to evaluate how accurately can the model extract the texts of these three types of textual elements from the documents. Specifically, we first compute the BLEU score for each extracted actor, action or constraint by the model according to the elements with the same type in the gold procedural graph~(e.g., for an extracted action, we find the most similar action in the gold procedural graph and compute the extracted action's BLEU score based on the most similar action), thus calculating the precision scores of these extracted three types of textual elements. Then we do the same computation for all actors, actions and constraints in the gold procedural graph according to the extracted elements by the model to calculate the recall scores of all these three types of textual elements in the gold procedural graph. Then we can calculate the F1 scores based on the precision scores and recall scores. Note that we calculate the F1 scores of actors based on the best matched action pair~(e.g., we first find the most similar action in the gold procedural graph for an extracted action and then calculate the BLEU score of the extracted actor by comparing these two actions' actors). 

For gateway extraction, we adopt the standard F1 scores to evaluate whether the model can correctly use gateways to organize the non-sequential actions in the procedural graphs. Due to the fact that the gateways are meaningful only when paired with the corresponding elements in the graphs~\cite{von2015business, dumas2018fundamentals}, we consider an extracted gateway is correct only if its type and at least one of its paired element match those of the gold procedural graph~(e.g., an extracted exclusive gateway and an action connecting with it by flow match a pair of such exclusive gateway and action in the gold procedural graph, this extracted exclusive gateway is considered correctly extracted). And for soft evaluation of the gateway extraction, an action or constraint is considered matched with the gold procedural graph if its BLEU score $\geq 0.5$. 

For flow extraction, we measure the performance via a soft metric~\cite{tandon-etal-2020-dataset}, which computes the F1 scores based on the BLEU scores of associated textual elements. 
Specifically, a flow is considered correctly extracted only if its type and connected elements match those of the gold procedural graph. For condition flow, we compute the BLEU score between its condition and the gold label's condition if another condition flow in the gold procedural graph can be matched.

\subsection{Details of Improvement Strategies}
\label{app:ProposedMethod}

\paragraph{Condition Verifier}

The designed templates used to provide feedback to the model are shown in Figure~\ref{fig:verifier_or} and Figure~\ref{fig:verifier_xor}. 
We feed the generated feedback based on the designed templates and extraction results to the model, so that the model can refine its extraction according to our provided feedback. 

\paragraph{Parallel Verifier}
We use the off-the-shelf parsing tool to parse the syntactic structure of the text and obtain the predicate and object corresponding to each action. 
The designed templates used to provide feedback to the model are shown in Figure~\ref{fig:verifier_parallel} and Figure~\ref{fig:verifier_sequential}.  

\begin{figure}[t]
    \centering
    \subfigure[The document to be extracted]{\includegraphics[width=\linewidth]{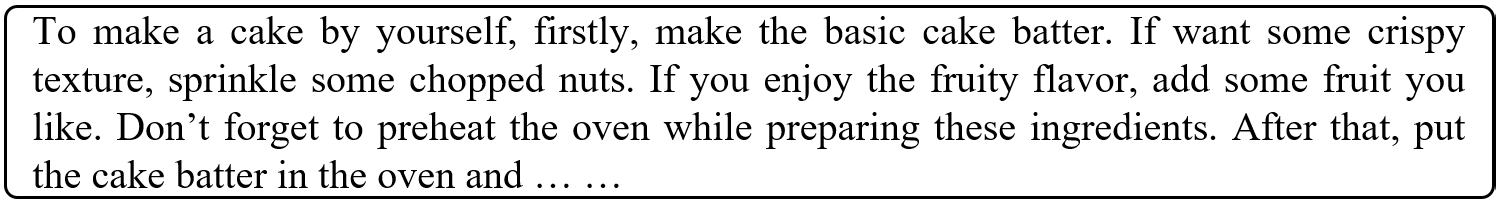}\label{fig:Case_text}}

    \vspace{0cm}
    \subfigure[Screenshot of the procedural graph extracted by PET]{\includegraphics[width=\linewidth]{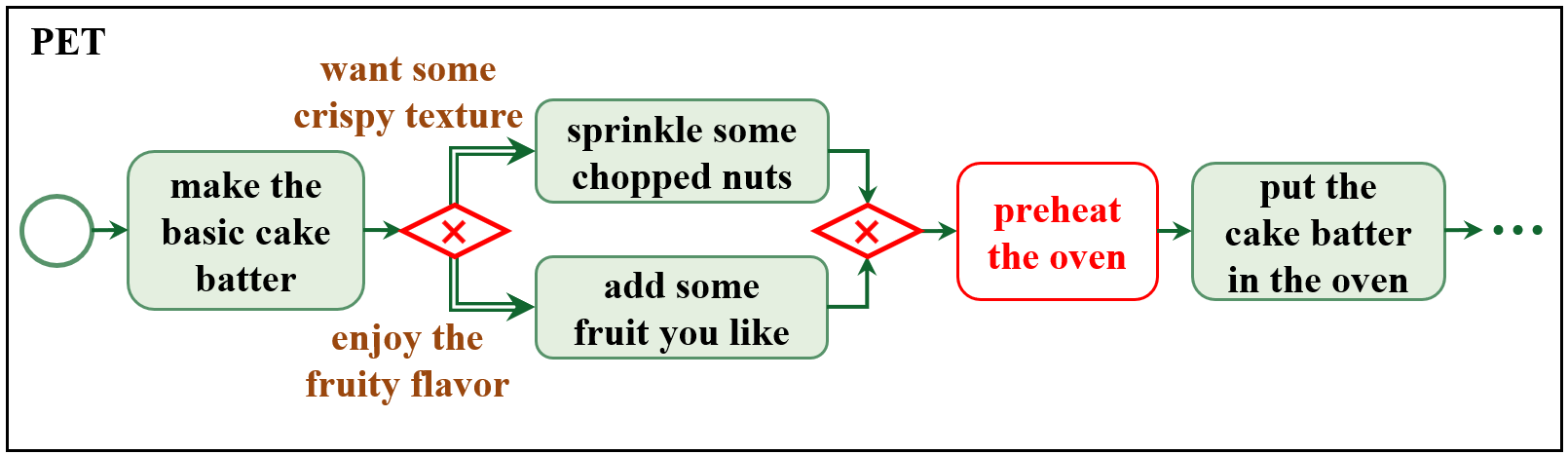}\label{fig:Case_pet}}

    \vspace{0cm}
    \subfigure[Screenshot of the procedural graph extracted by fine-tuned Llama2]{\includegraphics[width=\linewidth]{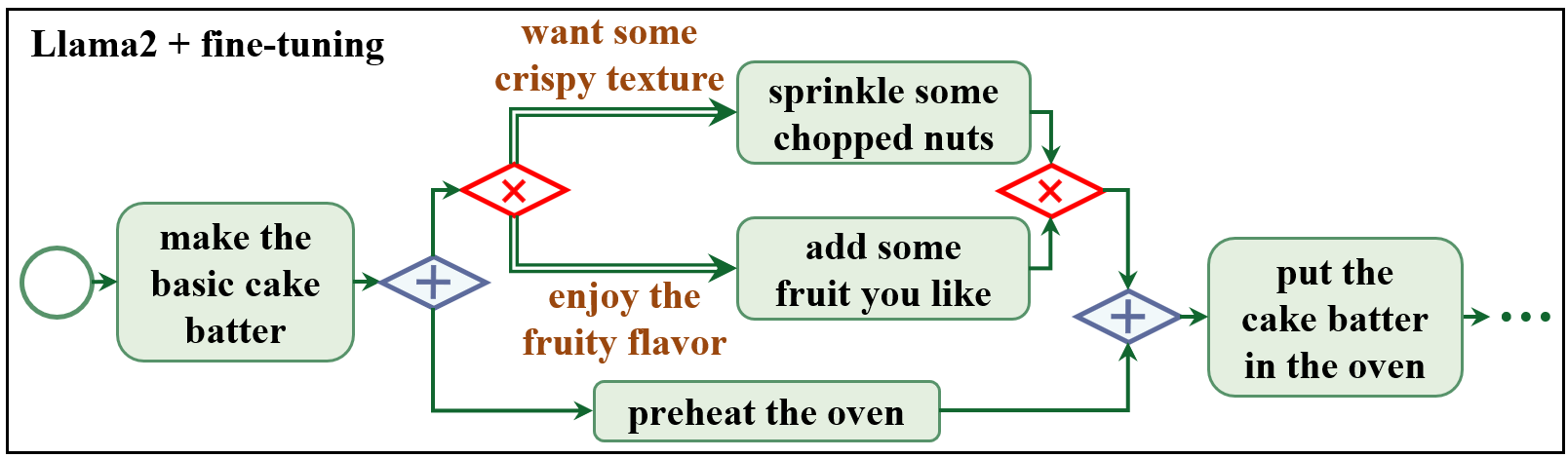}\label{fig:Case_tuning}}

    \vspace{0cm}
    \subfigure[Screenshot of the procedural graph extracted by fine-tuned Llama2 with self-refine strategy]{\includegraphics[width=\linewidth]{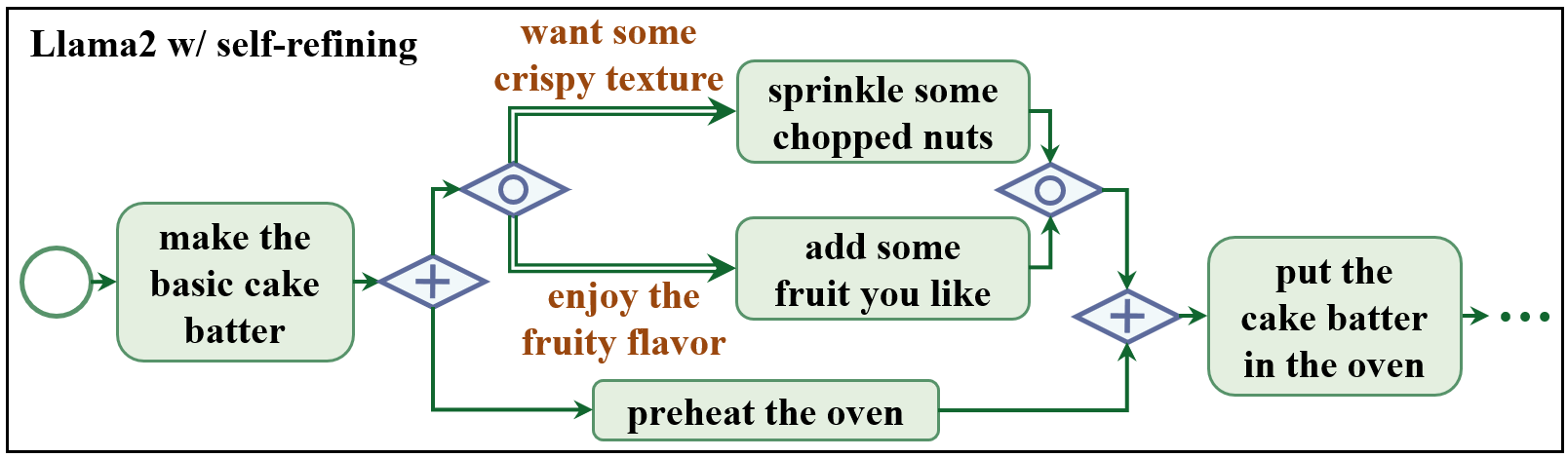}\label{fig:Case_self}}

    \caption{Illustration of case study.
    }
    \label{fig:case_study}
\end{figure}

\subsection{Case Study}
As shown in Figure~\ref{fig:case_study}, PET~\ref{fig:Case_pet} fails to handle the parallel actions in the document~\ref{fig:Case_text} due to the lack of the ability to understand complex expressions, and it uses the wrong type of gateway to organize the non-sequential actions ``sprinkle some chopped nuts'' and ``add some fruit you like'' as it can only deal with partial types of gateways. Fine-tuned Llama2~\ref{fig:Case_tuning} successfully organizes the parallel execution of actions in the document, but also uses the wrong gateway to organize the non-sequential actions ``sprinkle some chopped nuts'' and ``add some fruit you like''. With the help of our designed verifier~\ref{fig:Case_self}, fine-tuned Llama2 with self-refine strategy correct the wrong gateway into inclusive gateway through the feedback provided by the verifier, as there exists no conflict between the conditions ``want some crispy texture'' and ``enjoy the fruity flavor''. This demonstrates the effectiveness of our proposed strategy by paying extra attention to non-sequential logic prediction.

\begin{figure*}[!th]
    \centering
    \subfigure[Template for inclusive gateway refinement]{\includegraphics[width=1\textwidth]{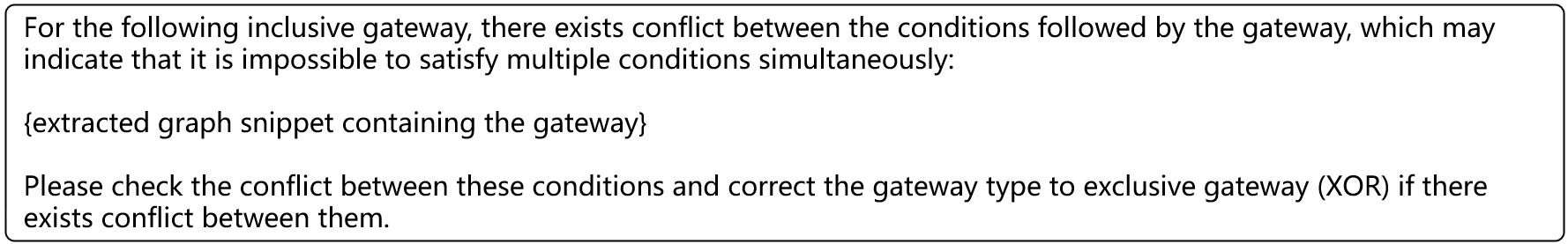}\label{fig:verifier_or}}

    \vspace{0cm}
    \subfigure[Template for exclusive gateway refinement]{\includegraphics[width=1\textwidth]{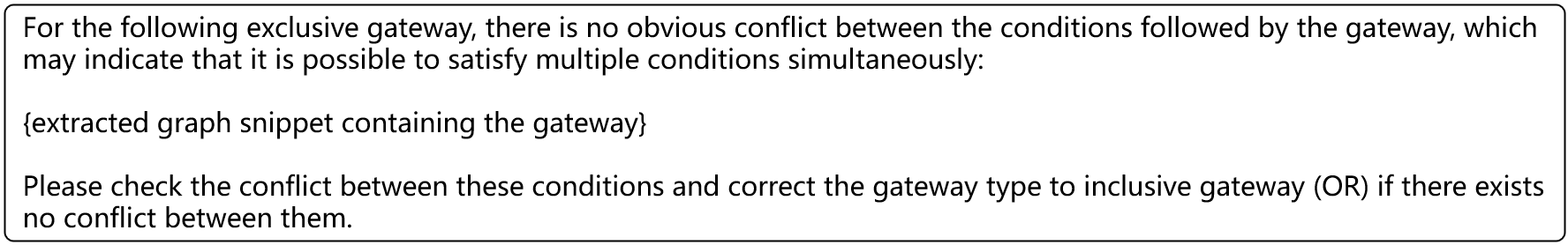}\label{fig:verifier_xor}}

    \vspace{0cm}
    \subfigure[Template for parallel gateway refinement]{\includegraphics[width=1\textwidth]{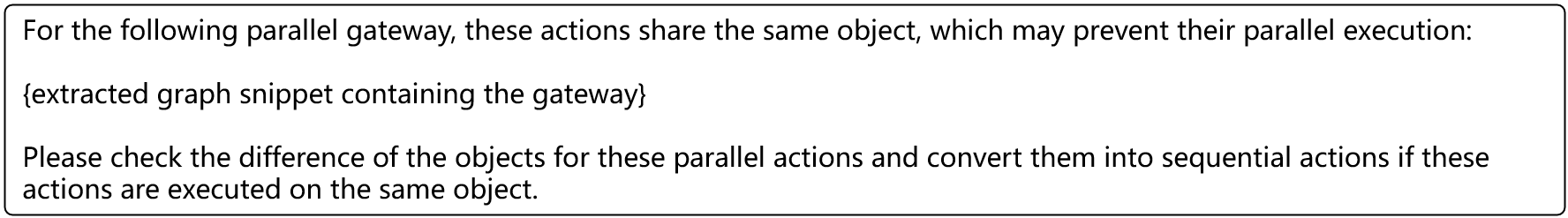}\label{fig:verifier_parallel}}

    \vspace{0cm}
    \subfigure[Template for sequential actions refinement]{\includegraphics[width=1\textwidth]{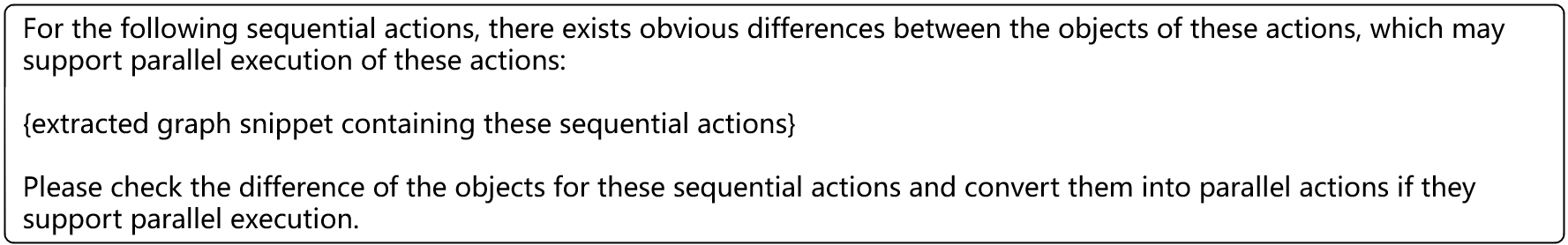}\label{fig:verifier_sequential}}

    \caption{Designed templates of our proposed verifiers.
    }
    \label{fig:verifier}
\end{figure*}

\begin{figure*}[t]
\ContinuedFloat
    \centering    
    \subfigure{\includegraphics[width=0.48\textwidth]{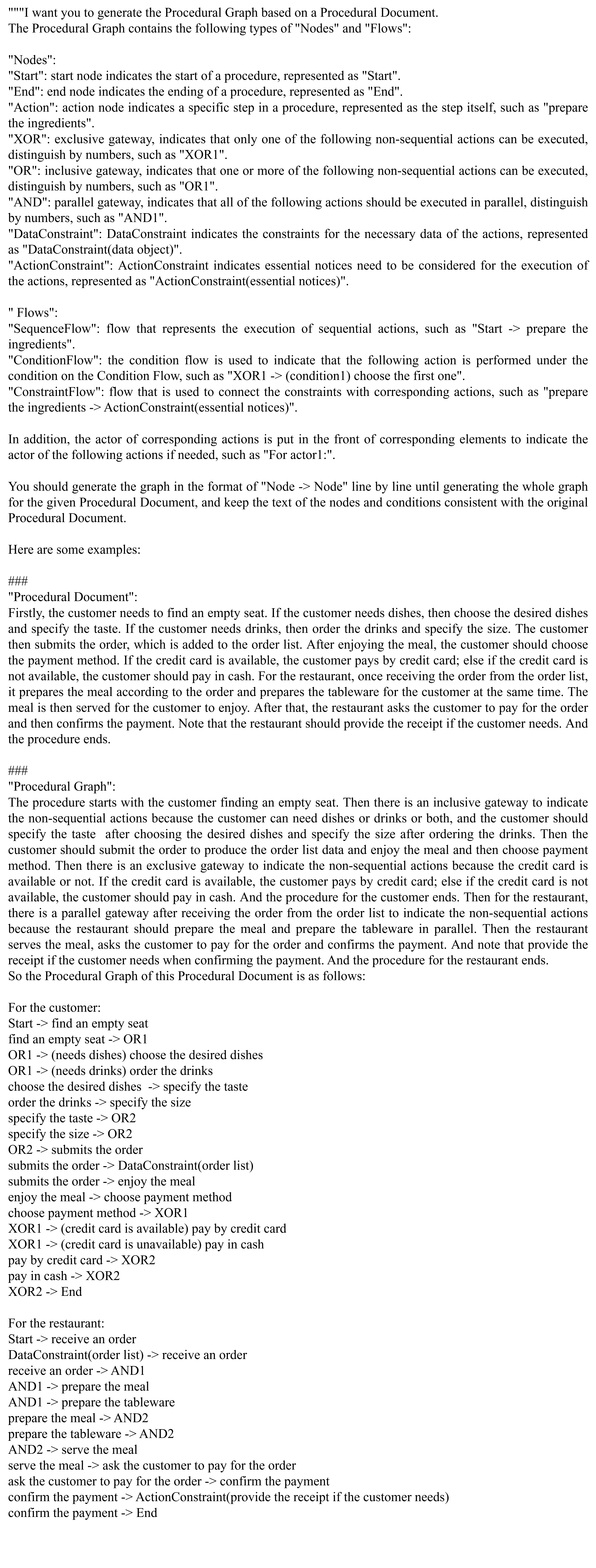}\label{fig:Prompt1}}\vspace{-0.3cm}    
    \subfigure{\includegraphics[width=0.48\textwidth]{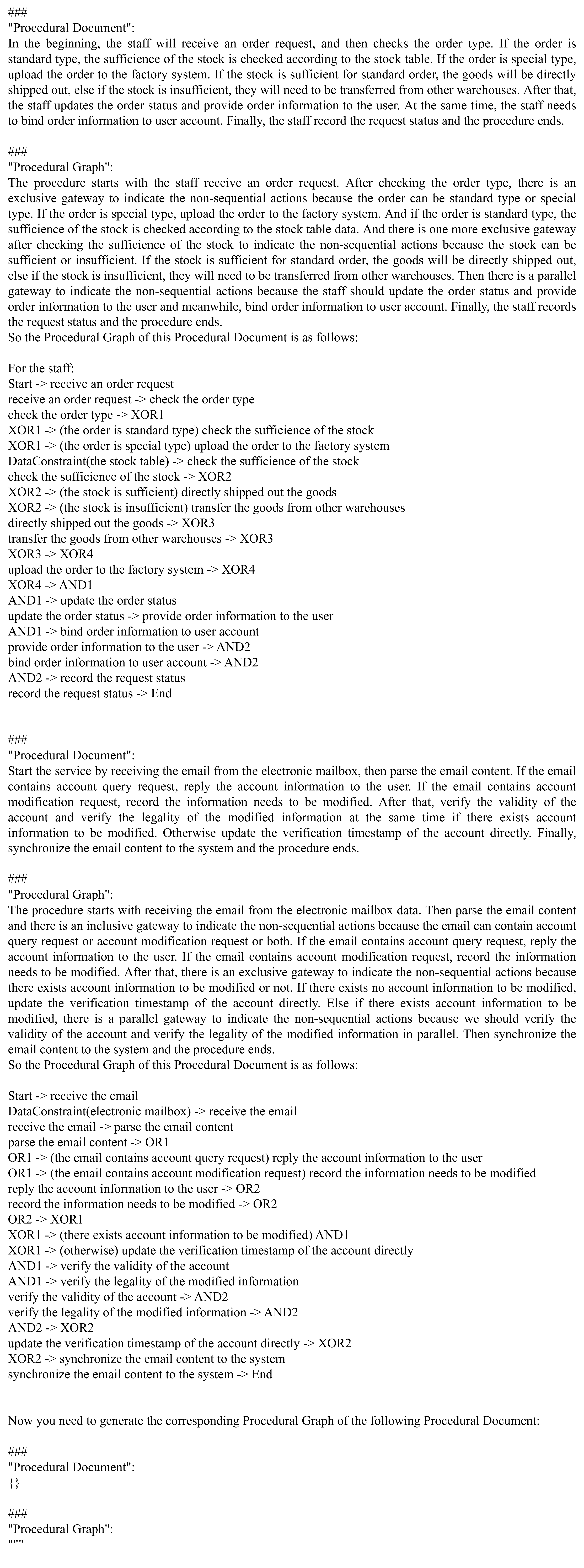}\label{fig:Prompt2}}    
    \caption{The adopted prompt consisting of elaborate instruction and three examples.
    }
    \label{fig:GenerationPrompt}
\end{figure*}





\end{document}